%% file: sn-bibliography.tex
\documentclass[default,iicol]{sn-jnl}

\usepackage{graphicx}%
\usepackage{multirow}%
\usepackage{amsmath,amssymb,amsfonts}%
\usepackage{amsthm}%
\usepackage{mathrsfs}%
\usepackage[title]{appendix}%
\usepackage[dvipsnames]{xcolor}
\usepackage{textcomp}%
\usepackage{manyfoot}%
\usepackage{booktabs}%
\usepackage{algorithm}%
\usepackage{algorithmicx}%
\usepackage{algpseudocode}%
\usepackage{listings}%
\usepackage{makecell, multirow, tabularx}
\usepackage{times}
\usepackage{epsfig}
\usepackage{tabularx}
\usepackage{arydshln}
\usepackage{bm}
\usepackage{nicefrac}
\usepackage{fixltx2e}
\usepackage{multirow}
\usepackage{mathtools}
\usepackage{algorithm}
\usepackage{algpseudocode}
\makeatletter
\@namedef{ver@everyshi.sty}{}
\makeatother
\usepackage{tikz}
\usetikzlibrary{patterns}
\input{plots}

\usepackage{booktabs}\let\cline\cmidrule
\usepackage{xspace}

\definecolor{amber}{rgb}{1.0, 0.75, 0.0}
\usepackage[capitalize]{cleveref}
\crefname{section}{Sec.}{Secs.}
\Crefname{section}{Section}{Sections}
\Crefname{table}{Table}{Tables}
\crefname{table}{Tab.}{Tabs.}
\renewcommand{\paragraph}[1]{\noindent\textbf{#1}}
\newcounter{daggerfootnote}

\theoremstyle{thmstyleone}%
\theoremstyle{thmstyletwo}%

\theoremstyle{thmstylethree}%

\begin{document}


\title[Three Things to Know about Deep Metric Learning]{Three Things to Know about Deep Metric Learning}


\author*{\fnm{Yash} \sur{Patel$^{1 *}$}} \email{yash.patel.research@gmail.com}

\author{\fnm{Giorgos} \sur{Tolias}}\email{toliageo@fel.cvut.cz}
\author{\fnm{Ji{\v{r}}{\'\i}} \sur{Matas}} \email{matas@fel.cvut.cz}

\affil[]{\orgdiv{VRG}, \orgname{FEE, Czech Technical University in Prague}, \country{Czechia}}

\input{abbrev}


\abstract{This paper addresses supervised deep metric learning for open-set image retrieval, focusing on three key aspects: the loss function, mixup regularization, and model initialization. In deep metric learning, optimizing the retrieval evaluation metric, recall@k, via gradient descent is desirable but challenging due to its non-differentiable nature. To overcome this, we propose a differentiable surrogate loss that is computed on large batches, nearly equivalent to the entire training set. This computationally intensive process is made feasible through an implementation that bypasses the GPU memory limitations. Additionally, we introduce an efficient mixup regularization technique that operates on pairwise scalar similarities, effectively increasing the batch size even further. The training process is further enhanced by initializing the vision encoder using foundational models, which are pre-trained on large-scale datasets. Through a systematic study of these components, we demonstrate that their synergy enables large models to nearly solve popular benchmarks.}


\maketitle
\footnotetext[1]{This work was done prior to joining Amazon.}

\input{0_intro}
\input{1_related}
\input{2_method}
\input{3_experiments}
\input{4_conclusions}

\footnotesize
\bibliographystyle{sn-nature}
\bibliography{sn-bibliography}

\end{document}

%% file: plots.tex
\usepackage{tikz}
\usetikzlibrary{arrows,shapes,calc,matrix,fit,backgrounds}
\usepackage{pgfplots}
\usepackage{pgfplotstable}
\pgfplotsset{compat=1.9}
\usepgfplotslibrary{fillbetween}

\usepackage{xstring}

\usepgfplotslibrary{external}

\IfBeginWith*{\jobname}{fig/extern/}{\finalcopy}{}

\tikzset{every mark/.append style={solid}}
\pgfplotsset{
	grid=both, width=\columnwidth, try min ticks=5,
	every axis/.append style={font=\scriptsize},
	every axis plot/.append style={thick,mark=none,mark size=1.2,tension=0.18},
	legend cell align=left, legend style={fill opacity=0.8},
}

\pgfplotsset{
	dash/.style={mark=o,dashed,opacity=0.7},
	dott/.style={mark=o,dotted,opacity=0.7},
}

%% file: abbrev.tex
\newcommand{\nn}[1]{\ensuremath{\text{NN}_{#1}}\xspace}
\def\l1{\ensuremath{\ell_1}\xspace}
\def\l2{\ensuremath{\ell_2}\xspace}

\def\roxf{$\mathcal{R}$Oxford\xspace}
\def\rox{$\mathcal{R}$Oxf\xspace}
\def\ro{$\mathcal{R}$O\xspace}
\def\rpar{$\mathcal{R}$Paris\xspace}
\def\rpa{$\mathcal{R}$Par\xspace}
\def\rp{$\mathcal{R}$P\xspace}
\def\rdis{$\mathcal{R}$1M\xspace}

\newcommand\resnet[3]{\ensuremath{\prescript{#2}{}{\mathtt{R}}{#1}_{\scriptscriptstyle #3}}\xspace}

\newcommand*\OK{\ding{51}}

\newenvironment{narrow}[1][1pt]
	{\setlength{\tabcolsep}{#1}}
	{\setlength{\tabcolsep}{6pt}}

\newcommand{\alert}[1]{{\color{red}{#1}}}
\newcommand{\gio}[1]{{\color{blue}{#1}}}
\newcommand{\replace}[2]{{\color{gray}{#1}}{\color{red}{#2}}}

\newcommand{\comment} [1]{{\color{orange} \Comment     #1}} 


\newcommand{\head}[1]{{\smallskip\noindent\bf #1}}
\newcommand{\equ}[1]{(\ref{equ:#1})\xspace}

\newcommand{\red}[1]{{\color{red}{#1}}}
\newcommand{\blue}[1]{{\color{blue}{#1}}}
\newcommand{\green}[1]{{\color{green}{#1}}}
\newcommand{\gray}[1]{{\color{gray}{#1}}}


\newcommand{\tran}{^\top}
\newcommand{\mtran}{^{-\top}}
\newcommand{\zcol}{\mathbf{0}}
\newcommand{\zrow}{\zcol\tran}

\newcommand{\ind}{\mathds{1}}
\newcommand{\expect}{\mathbb{E}}
\newcommand{\nat}{\mathbb{N}}
\newcommand{\zahl}{\mathbb{Z}}
\newcommand{\real}{\mathbb{R}}
\newcommand{\proj}{\mathbb{P}}
\newcommand{\prob}{\mathbf{Pr}}

\newcommand{\mif}{\textrm{if }}
\newcommand{\other}{\textrm{otherwise}}
\newcommand{\minimize}{\textrm{minimize }}
\newcommand{\maximize}{\textrm{maximize }}

\newcommand{\id}{\operatorname{id}}
\newcommand{\const}{\operatorname{const}}
\newcommand{\sgn}{\operatorname{sgn}}
\newcommand{\var}{\operatorname{Var}}
\newcommand{\mean}{\operatorname{mean}}
\newcommand{\trace}{\operatorname{tr}}
\newcommand{\diag}{\operatorname{diag}}
\newcommand{\vect}{\operatorname{vec}}
\newcommand{\cov}{\operatorname{cov}}

\newcommand{\softmax}{\operatorname{softmax}}
\newcommand{\clip}{\operatorname{clip}}

\newcommand{\defn}{\mathrel{:=}}
\newcommand{\peq}{\mathrel{+\!=}}
\newcommand{\meq}{\mathrel{-\!=}}

\newcommand{\floor}[1]{\left\lfloor{#1}\right\rfloor}
\newcommand{\ceil}[1]{\left\lceil{#1}\right\rceil}
\newcommand{\inner}[1]{\left\langle{#1}\right\rangle}
\newcommand{\norm}[1]{\left\|{#1}\right\|}
\newcommand{\frob}[1]{\norm{#1}_F}
\newcommand{\card}[1]{\left|{#1}\right|\xspace}
\newcommand{\diff}{\mathrm{d}}
\newcommand{\der}[3][]{\frac{d^{#1}#2}{d#3^{#1}}}
\newcommand{\pder}[3][]{\frac{\partial^{#1}{#2}}{\partial{#3^{#1}}}}
\newcommand{\ipder}[3][]{\partial^{#1}{#2}/\partial{#3^{#1}}}
\newcommand{\dder}[3]{\frac{\partial^2{#1}}{\partial{#2}\partial{#3}}}

\newcommand{\wb}[1]{\overline{#1}}
\newcommand{\wt}[1]{\widetilde{#1}}

\def\nsp{\hspace{-3pt}}
\def\zsp{\hspace{0pt}}
\def\xssp{\hspace{1pt}}
\def\ssp{\hspace{3pt}}
\def\msp{\hspace{6pt}}
\def\lsp{\hspace{12pt}}
\def\xlsp{\hspace{20pt}}

\newcommand{\cA}{\mathcal{A}}
\newcommand{\cB}{\mathcal{B}}
\newcommand{\cC}{\mathcal{C}}
\newcommand{\cD}{\mathcal{D}}
\newcommand{\cE}{\mathcal{E}}
\newcommand{\cF}{\mathcal{F}}
\newcommand{\cG}{\mathcal{G}}
\newcommand{\cH}{\mathcal{H}}
\newcommand{\cI}{\mathcal{I}}
\newcommand{\cJ}{\mathcal{J}}
\newcommand{\cK}{\mathcal{K}}
\newcommand{\cL}{\mathcal{L}}
\newcommand{\cM}{\mathcal{M}}
\newcommand{\cN}{\mathcal{N}}
\newcommand{\cO}{\mathcal{O}}
\newcommand{\cP}{\mathcal{P}}
\newcommand{\cQ}{\mathcal{Q}}
\newcommand{\cR}{\mathcal{R}}
\newcommand{\cS}{\mathcal{S}}
\newcommand{\cT}{\mathcal{T}}
\newcommand{\cU}{\mathcal{U}}
\newcommand{\cV}{\mathcal{V}}
\newcommand{\cW}{\mathcal{W}}
\newcommand{\cX}{\mathcal{X}}
\newcommand{\cY}{\mathcal{Y}}
\newcommand{\cZ}{\mathcal{Z}}

\newcommand{\vA}{\mathbf{A}}
\newcommand{\vB}{\mathbf{B}}
\newcommand{\vC}{\mathbf{C}}
\newcommand{\vD}{\mathbf{D}}
\newcommand{\vE}{\mathbf{E}}
\newcommand{\vF}{\mathbf{F}}
\newcommand{\vG}{\mathbf{G}}
\newcommand{\vH}{\mathbf{H}}
\newcommand{\vI}{\mathbf{I}}
\newcommand{\vJ}{\mathbf{J}}
\newcommand{\vK}{\mathbf{K}}
\newcommand{\vL}{\mathbf{L}}
\newcommand{\vM}{\mathbf{M}}
\newcommand{\vN}{\mathbf{N}}
\newcommand{\vO}{\mathbf{O}}
\newcommand{\vP}{\mathbf{P}}
\newcommand{\vQ}{\mathbf{Q}}
\newcommand{\vR}{\mathbf{R}}
\newcommand{\vS}{\mathbf{S}}
\newcommand{\vT}{\mathbf{T}}
\newcommand{\vU}{\mathbf{U}}
\newcommand{\vV}{\mathbf{V}}
\newcommand{\vW}{\mathbf{W}}
\newcommand{\vX}{\mathbf{X}}
\newcommand{\vY}{\mathbf{Y}}
\newcommand{\vZ}{\mathbf{Z}}

\newcommand{\va}{\mathbf{a}}
\newcommand{\vb}{\mathbf{b}}
\newcommand{\vc}{\mathbf{c}}
\newcommand{\vd}{\mathbf{d}}
\newcommand{\ve}{\mathbf{e}}
\newcommand{\vf}{\mathbf{f}}
\newcommand{\vg}{\mathbf{g}}
\newcommand{\vh}{\mathbf{h}}
\newcommand{\vi}{\mathbf{i}}
\newcommand{\vj}{\mathbf{j}}
\newcommand{\vk}{\mathbf{k}}
\newcommand{\vl}{\mathbf{l}}
\newcommand{\vm}{\mathbf{m}}
\newcommand{\vn}{\mathbf{n}}
\newcommand{\vo}{\mathbf{o}}
\newcommand{\vp}{\mathbf{p}}
\newcommand{\vq}{\mathbf{q}}
\newcommand{\vr}{\mathbf{r}}
\newcommand{\Vs}{\mathbf{s}}
\newcommand{\vt}{\mathbf{t}}
\newcommand{\vu}{\mathbf{u}}
\newcommand{\vv}{\mathbf{v}}
\newcommand{\vw}{\mathbf{w}}
\newcommand{\vx}{\mathbf{x}}
\newcommand{\vy}{\mathbf{y}}
\newcommand{\vz}{\mathbf{z}}

\newcommand{\vone}{\mathbf{1}}
\newcommand{\vzero}{\mathbf{0}}

\newcommand{\valpha}{{\boldsymbol{\alpha}}}
\newcommand{\vbeta}{{\boldsymbol{\beta}}}
\newcommand{\vgamma}{{\boldsymbol{\gamma}}}
\newcommand{\vdelta}{{\boldsymbol{\delta}}}
\newcommand{\vepsilon}{{\boldsymbol{\epsilon}}}
\newcommand{\vzeta}{{\boldsymbol{\zeta}}}
\newcommand{\veta}{{\boldsymbol{\eta}}}
\newcommand{\vtheta}{{\boldsymbol{\theta}}}
\newcommand{\viota}{{\boldsymbol{\iota}}}
\newcommand{\vkappa}{{\boldsymbol{\kappa}}}
\newcommand{\vlambda}{{\boldsymbol{\lambda}}}
\newcommand{\vmu}{{\boldsymbol{\mu}}}
\newcommand{\vnu}{{\boldsymbol{\nu}}}
\newcommand{\vxi}{{\boldsymbol{\xi}}}
\newcommand{\vomikron}{{\boldsymbol{\omikron}}}
\newcommand{\vpi}{{\boldsymbol{\pi}}}
\newcommand{\vrho}{{\boldsymbol{\rho}}}
\newcommand{\vsigma}{{\boldsymbol{\sigma}}}
\newcommand{\vtau}{{\boldsymbol{\tau}}}
\newcommand{\vupsilon}{{\boldsymbol{\upsilon}}}
\newcommand{\vphi}{{\boldsymbol{\phi}}}
\newcommand{\vchi}{{\boldsymbol{\chi}}}
\newcommand{\vpsi}{{\boldsymbol{\psi}}}
\newcommand{\vomega}{{\boldsymbol{\omega}}}

\newcommand{\rLambda}{\mathrm{\Lambda}}
\newcommand{\rSigma}{\mathrm{\Sigma}}

\makeatletter
\DeclareRobustCommand\onedot{\futurelet\@let@token\@onedot}
\def\@onedot{\ifx\@let@token.\else.\null\fi\xspace}
\def\eg{\emph{e.g}\onedot} 
\def\Eg{\emph{E.g}\onedot}
\def\ie{\emph{i.e}\onedot} 
\def\Ie{\emph{I.e}\onedot}
\def\cf{\emph{cf}\onedot} 
\def\Cf{\emph{C.f}\onedot}
\def\etc{\emph{etc}\onedot} 
\def\vs{\emph{vs}\onedot}
\def\wrt{w.r.t\onedot} 
\def\dof{d.o.f\onedot}
\def\etal{\emph{et al}\onedot}
\makeatother

%% file: 0_intro.tex
\section{Introduction}
\label{sec:introduction}

Deep metric learning (DML) is a representation learning task with deep models, typically aiming to retrieval or classification with non-parametric, \ie nearest neighbor, classifiers. 
For retrieval, the task is considered open-set with evaluation performed on classes that are not seen during training. 
Therefore, the ability to generalize well is crucial to achieve good performance, measured using established information retrieval metrics, such as mean Average Precision (mAP) or recall@k.
Due to the common limitation of having limited training data on particular domains, it's essential to initialize the model properly rather than training from scratch. 
This work focuses on DML for images, aiming to  bridge the gap between learning and evaluation objectives, while exploring the significance of model initialization.

\begin{figure*}[t]
\vspace{10pt}
    \centering
    \includegraphics[width=\textwidth]{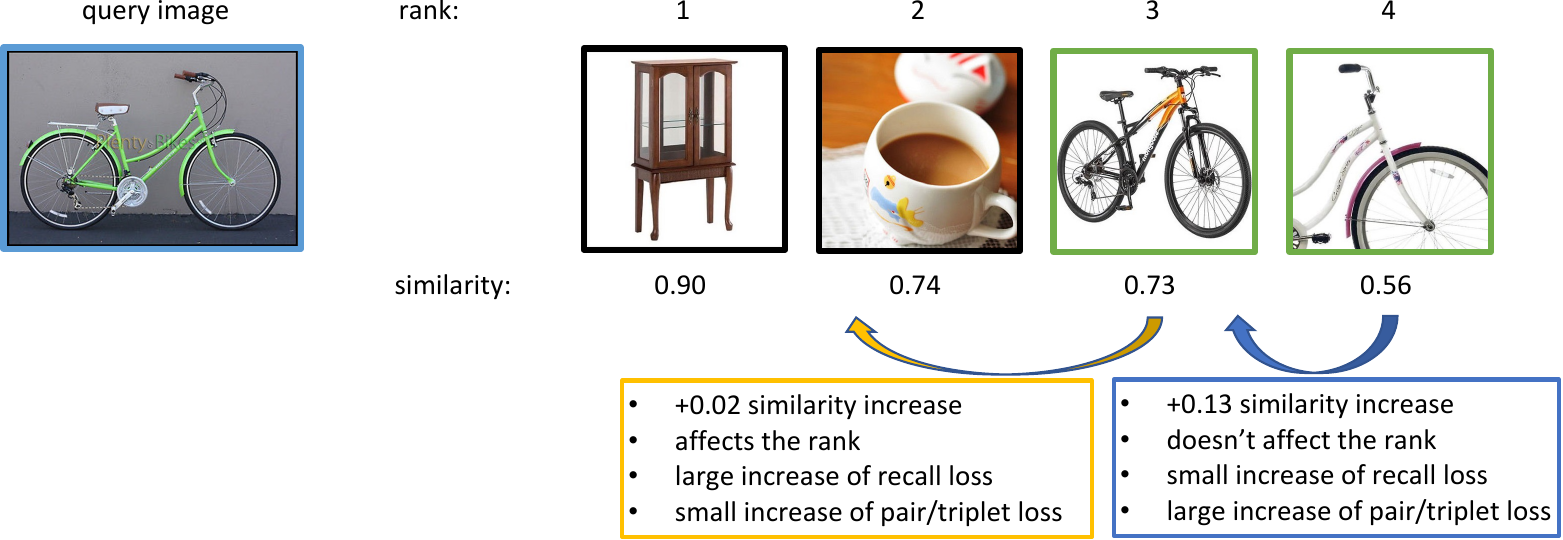}
\vspace{5pt}    
        \caption{Example of a training batch with a query image, two positive and two negative images. 
        Despite not positively affecting the image ranking, large similarity changes may correspond to large loss with pair-based or triplet-based loss functions, such as contrastive or triplet loss. The proposed recall loss reflects the test-time evaluation metric and focuses on similarity changes that have a positive impact on the ranks. 
    \label{fig:motivation_retrieval}
    }
\vspace{10pt}    
\end{figure*}

\begin{figure}[t]
\begin{center}
\includegraphics[width=0.48\textwidth]{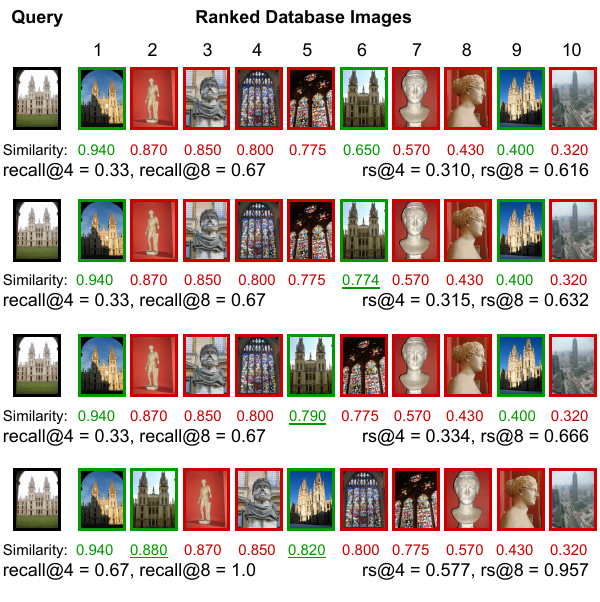}
\vspace{-5pt}
\caption{A comparison between recall@k and rs@k, the proposed differentiable recall@k surrogate. Examples show a query, the ranked database images sorted according to the similarity and the corresponding values for recall@k and rs@k and their dependence on similarity score change. Note that the values of recall@k and rs@k are close. Changes to similarity and ranking in some cases may not affect the original recall@k but can affect the surrogate, with the latter having a more significant impact than the former. Similarity values of all negatives are fixed for ease of understanding. The similarity values of the positives that were changed in rows 2, 3 and 4 are \underline{underlined}.
\label{fig:teaser}}
\end{center}
\end{figure}

Minimization of a loss that is a function of the test-time evaluation metric has shown to be beneficial in deep learning for numerous computer vision and natural language processing tasks. 
Examples include intersection-over-union as a loss that boosts performance for object detection~\cite{yjw+16,rig+19} and semantic segmentation~\cite{nsb+18}, and structural similarity~\cite{mae+18}, peak signal-to-noise ratio~\cite{bms+18} and perceptual~\cite{pam+21} as reconstruction losses for image compression that give better results according to the respective evaluation metrics. 
Training deep networks via gradient descent on the evaluation metric is not possible when the metric is non-differentiable. 
To address this, existing DML methods  use proxy loss functions like contrastive~\cite{hcl06}, triplet~\cite{skp+15}, and margin~\cite{wms+17}.
For a distance measure, these loss functions encourage pulling together samples from the same class while pushing apart samples from different classes. 
As differentiable functions, they provide a workaround, which empirically leads to a reasonable performance, but may not align well with the evaluation metric as shown in Figure \ref{fig:motivation_retrieval}. 

We propose a hand-designed smooth approximation of the recall@k metric as a surrogate function for training, and a mixup technique that efficiently operates in the space of image-to-image similarities.
Figure \ref{fig:teaser} shows a comparison between the evaluation metric and the proposed loss.
Generic methods for training with non-differentiable losses, such as actor-critic~\cite{bbx+17} and learning surrogates~\cite{phm+20} are not directly applicable to recall@k. This is due to the fact that these methods are limited to decomposable functions, where a per-example performance measure is available. Such an attempt is made by Engilberge~\etal~\cite{ecp+19}, where an LSTM learns sorting-based metrics, but is not adapted in consequent work due to slow training.

In a variety of DML benchmarks~\cite{vms+18,vms+18,ohb16,ksd+13}, it is common practice  to initialize the model with the weights from ImageNet~\cite{dsl+09} classification-based pre-training. As compared to initializing the model from scratch due to limited training data availability for the down-stream task. The recent literature on learning visual representations includes significant efforts in large-scale pre-training, conducted either fully supervised~\cite{dbk+21}, weakly~\cite{sga+22,rdk+23,rkh+21} supervised, or self-supervised~\cite{odm+23,ctm+21,cmm+20}. These pre-trained models aim for effective transfer learning, whether kept frozen or fine-tuned. However, their impact on DML remains largely unexplored. This work investigates the effects of initializing models with the outcomes of widely known large-scale pre-training methods.

\noindent
The list of contributions of this work, which is an extension of our prior work~\cite{ptm22}, are outlined below.
\begin{itemize}
\setlength\itemsep{6pt}
\item A new loss function is proposed, which is a surrogate of recall at top $k$. To mimic test-time settings, we train with very large batch sizes borrowing implementation tricks from prior work~\cite{rar+19}. 
\item We further enlarge the batch size through a new mixup-based regularization technique that is computationally efficient and that virtually enlarges the batch, without creating the mixed embeddings. Its efficiency is obtained by operating on the very last stage of similarity estimation, \ie scalar similarities are mixed.
\item The impact of different pre-trainings used as model initialization is explored by included widely known methods, namely CLIP~\cite{rkh+21}, DiHT~\cite{rdk+23}, DINOv2~\cite{odm+23}, SWAG~\cite{sga+22} and training on ImageNet-21k~\cite{dbk+21}.
\item We use a fair validation protocol to perform hyper-parameter tuning, which is not the standard practice in the literature.
\item We achieve nearly perfect results on several DML benchmarks using large backbones, strong model initialization and the proposed loss.
\end{itemize}

With the above mentioned contributions, we demonstrate it is possible to achieve recall@1 scores of $90.0\%$ on iNaturalist~\cite{vms+18}, $90.8\%$ on Stanford-Online-Products~\cite{ohb16}, and $97.2\%$ on Cars196~\cite{ksd+13} datasets. Furthermore we archive $97.6\%$ Recall@16 on iNaturalist, $97.7\%$ Recall@10 on SOP, and $99.3\%$ Recall@8 Cars196, which signifies that under certain settings those benchmarks are nearly solved. 
We hope that the diverse experimentation done in this paper with different architectures and initializations will provide guidance and baselines for future exploration in the field.

%% file: 1_related.tex
\section{Related work}
\label{sec:related_work}

In this section, the related work is reviewed for two different families of deep metric learning approaches regarding the type of loss that is optimized, namely classification losses and pairwise losses. 
Given an embedding network that maps input images to a high dimensional space, in the former, the loss is a function of the embedding and the corresponding category label of a single image, while in the latter, the loss is a function of the distance, or similarity, between two embeddings and the corresponding pairwise label. 
Prior work for mixup~\cite{zcd+17} techniques related to embedding learning is reviewed too.

\paragraph{Classification losses.} The work of Zhai and Wu~\cite{zw18} supports that the standard classification loss, \ie cross-entropy (CE) loss is a strong approach for deep metric learning. Their finding is supported by the use of layer normalization and class-balanced sampling. In the domain of metric learning for faces, several different classification losses are proposed, such as SphereFace~\cite{lwy+17}, CosFace~\cite{wwz+18} and ArcFace~\cite{dgx+19}, where contributions are in the spirit of large margin classification. Despite the specificity of the domain, such losses are applicable beyond faces. Another variant is the Neighborhood Component Analysis (NCA) loss that is used in the work of Movshovitz-Attias~\etal~\cite{mtl+17}, which is later improved~\cite{tdt20} by temperature-based scaling and faster update of the class prototype vectors, also called proxies in their work. The restriction of a single prototype vector per class is dropped by Qian~\etal~\cite{qss+19} who stores multiple representatives per category. 
Instead of jointly processing the example to all class vectors similarities, Kim \etal~\cite{kim2020proxy} jointly normalize the class vector to all batch examples similarities. 
Their interpretation is that such a formulation allows examples to interact with each other, indirectly resembling pairwise losses.

Classification losses, in contrast to pairwise losses, perform the optimization independently per image. An exception is the work of Elezi~\etal~\cite{evt+20} where a similarity propagation module captures group interactions within the batch. Then, cross-entropy loss is used, which now comes with significant improvements by taking into account such interactions. This is later improved~\cite{sel21,kpm+23} by replacing the propagation module with an attention model.
The relation between CE loss and some of the widely used pairwise losses is studied from a mutual information point of view~\cite{brz+20}. CE loss is viewed as approximate bound-optimization for minimizing pairwise losses; CE maximizes mutual information, and so do these pairwise losses, which are reviewed in the following.

\paragraph{Pairwise losses.} The first pairwise loss introduced is contrastive loss~\cite{hcl06}, where embeddings of relevant pairs are pushed as close as possible, while those of non-relevant ones are pushed far enough. Since the target task is typically a ranking one, the triplet loss~\cite{skp+15}, a popular and widely used loss, improves that by forming training triplets in the form of anchor, positive and negative examples. The loss is a function of the difference between anchor-to-positive and anchor-to-negative distances and is zero if such a difference is large enough, therefore satisfying the objectives of a ranking task for this triplet. Optimization over all pairs or triplets is not tractable and is observed to be sub-optimal~\cite{wms+17}. As a result, a lot of attention is paid to finding informative pairs and triplets~\cite{mbl20,rms+20,sxj+15,sohn16}, which typically includes heuristics. Several other losses are suggested in the literature~\cite{wms+17,wzw+17,sxj+15,wlf+23} and are added to the long list of hand-designed proxy losses which target to learn embeddings that transfer well to a ranking or a similar task.

A few cases follow a principled approach for obtaining a loss that is appropriate for ranking tasks. This is the case with the work of Ustinova~\etal~\cite{ul16} where the goal is to minimize the probability that the similarity between embeddings of a non-relevant pair is larger than that of a relevant one. This probability is approximated by the quantization of the range of possible similarities and the histogram loss, which is estimated within a single batch. Their work dispenses with the need for any kind of sampling for mini-batch construction. 
Another example of optimizing principled metrics is based on mutual information~\cite{Kemertas_2020_CVPR}. Another case in the same direction focuses on optimizing AP, which is a standard retrieval evaluation metric. A smooth approximation of it is often used in the literature~\cite{rmp+20,hls18,rar+19}, while the work of Brown~\etal~\cite{bxk+20} is the closest to ours. In their work, the step function is approximated with a sigmoid, while Ramzi \etal~\cite{ramzi2021robust} propose a piecewise alternative after observing that the function should rather not be symmetric. 
In their follow-up work~\cite{ramzi2022hierarchical} access to a class hierarchy is assumed and taken into account within the loss function.
Both these contributions of Ramzi and colleagues, are orthogonal to, and theoretically can be combined with, ours in this work.
While using AP-based losses, a large batch size is crucial, which meets the limitations set by the hardware. Such limitations are overcome in the work of Revaud~\etal~\cite{rar+19} who uses a batch of $4,000$ high-resolution images.

\paragraph{Embedding mixup.} Manifold mixup~\cite{vlb+19}, which involves mixing~\cite{zcd+17} intermediate representations and labels of two examples, has demonstrated to improve generalizability for supervised learning by encouraging smoother decision boundaries. Such techniques are investigated for embedding learning and image retrieval by mixing the embedding of two examples. Duan~\etal~\cite{dzl+18} uses adversarial training to synthesize additional negative samples from the observed negatives. Kalantidis ~\etal~\cite{ksp+20} synthesize hard-negatives for contrastive self-supervised learning by mixing the embedding of the two hardest negatives and also mixing them with the query itself. Zheng~\etal~\cite{zcl+19} uses a linear interpolation between the embeddings to manipulate the hardness levels. In the work of Gu ~\etal~\cite{gk+20}, two embedding vectors from the same class are used to generate symmetrical synthetic examples and hard-negative mining is performed within the set of original and the synthetic examples. This is further extended to proxy-based losses, where the embedding of examples from different classes and labels is mixed to generate synthetic proxies~\cite{gkk+21}. Linearly interpolating labels entails the risk of generating false negatives if the interpolation factor is close to $0$ or $1$. Such limitations are overcome in the work of Venkataramanan ~\etal~\cite{vpa+21}, which generalizes mixing examples from different classes for pairwise loss functions. The proposed {\em SiMix} approach differs from the aforementioned techniques as it operates on the similarity scores instead of the embedding vectors, does not require training an additional model, making it computationally efficient. Furthermore, unlike the existing mixup techniques, it uses a synthetic sample in the roles of a query, positive and negative example. 

%% file: 2_method.tex
\section{Method}
\label{sec:method}

We present the task, the relevant background and the proposed approaches.

\subsection{Background}

\paragraph{Task.} We are given a query example $q\in \cX$ and a collection of examples $\Omega \subset \cX$, also called database, where $\cX$ is the space of all images. The set of database examples that are positive or negative to the query are denoted by $P_q$ and $N_q$, respectively, with $\Omega=P_q \cup N_q$. Ground-truth information for the positive and negative sets per query is obtained according to discrete class labels per example, \ie if two examples come from the same class, then they are considered positive to each other, otherwise negative. This is the case for all (training or testing) databases used in this work. Terms example and image are used interchangeably in the following text. In image retrieval, all database images are ranked according to similarity to the query $q$, and the goal is to rank positive examples before negative ones.

\paragraph{Deep image embeddings.} 
Image embeddings, otherwise called descriptors, are generated by function $f_{\theta}: \cX \rightarrow \real^d$. In this work, function $f_\theta$ is a convolutional neural network~\cite{hzr+16} or a vision transformer~\cite{dbk+21} mapping input images of any size or aspect ratio to an $L_{2}$-normalized $d$-dimensional embedding. 
Embedding for image $x$ is denoted by $\vx = f_\theta(x)$. Given a pre-trained backbone, linear projection layers are used to map the backbone's output to the desired dimension $d$, which are part of the parameter set $\theta$. Parameter set $\theta$ of the network is learned (fine-tuned) during the training. Similarity between a query $q$ and a database image $x$ is computed by the dot product of the corresponding embeddings and is denoted by $s(q,x) = \vq^\top \vx$, also denoted as $s_{qx}$ for brevity.

\paragraph{Evaluation metric.} Recall@k is one of the standard metrics to evaluate image retrieval methods. For query $q$, it is defined as a ratio of the number of relevant (positive) examples within the top-k ranked examples to the total number of relevant  examples for $q$ given by $|P_q|$.
It is denoted by $R_\Omega^k(q)$ when computed for query $q$ and database $\Omega$ and can be expressed as

\begin{equation}
    R_\Omega^k(q) = \frac{\sum\limits_{x \in P_q} H (k - r_\Omega(q,x))}{|P_q|},
    \label{equ:recall}
\end{equation}
where $r_\Omega(q,x)$ is the rank of example $x$ when all database examples in $\Omega$ are ranked according to similarity to query $q$. Function $H(.)$ is the Heaviside step function, which is equal to 0 for negative values, otherwise equal to 1. The rank of example $x$ is computed by

\begin{equation}
    r_\Omega(q,x) = 1 + \sum\limits_{z \in \Omega, z \neq x} H(s_{qz} - s_{qx}),
    \label{equ:rank}
\end{equation}

Therefore, \equ{recall} can now be expressed as
\begin{equation}
R_\Omega^k(q) = \frac{\sum\limits_{x \in P_q} H (k  - 1 - \sum\limits_{z \in \Omega, z \neq x} H(s_{qz} - s_{qx}) )}{|P_q|}.
\label{equ:recall_full}
\end{equation}

\subsection{Recall@k surrogate loss} 
\begin{figure}[t]
\begin{center}
\input{fig_sigmoid}
\hspace{-10pt}
\caption{The two sigmoid functions which replace the Heaviside step function for counting the positive examples in the short-list of size $k$ (left) and for estimating the rank of examples (right).
\label{fig:sigmoids}
}
\end{center}
\end{figure}
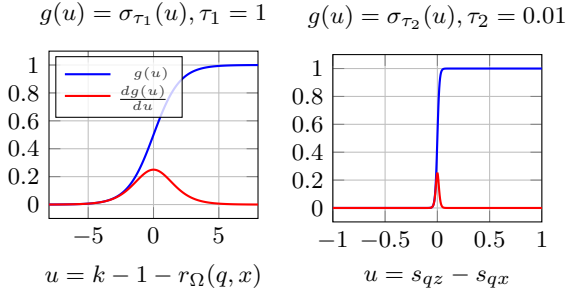

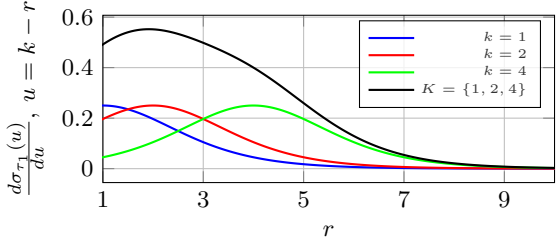
\begin{figure}[t]
\hspace{-5pt}
\begin{center}
\input{fig_deriv}
\hspace{-15pt}
\caption{Gradient magnitude of the sigmoid used to count the positive examples in the short-list of size $k$ versus the rank $r$ (equal to $r_\Omega(q,x)$, see \equ{rank}) of a positive example $x$. It shows how much a positive example is pushed towards lower ranks depending on its current rank. In the case of multiple values for $k$, the total gradient is equivalent to the sum of the separate ones.
\label{fig:deriv}
}
\end{center}
\end{figure}

The computation of recall in \equ{recall_full} involves the use of the Heaviside step function. The gradient of the Heaviside step function is a Dirac delta function. Hence, direct optimization of recall with back-propagation is not feasible. A common smooth approximation of the Heaviside step function is provided by the logistic function~\cite{km+15,ikm+15,ikm+17}, a common sigmoid function $\sigma_{\tau}: \real \rightarrow \real$ controlled by temperature $\tau$, which is given by
\begin{equation}
\sigma_\tau(u) = \frac{1}{1+e^{-\frac{u}{\tau}}},
\end{equation}
where large (small) temperature value leads to worse (better) approximation and denser (sparser) gradient. This approximation is common in the machine learning literature for several tasks~\cite{sh+09,gls+16,mmt+17} and also appears in the approximation of the Average Precision evaluation metric~\cite{bxk+20}, which is used for the same task as ours. By replacing the step function with the sigmoid function, a smooth approximation of recall is obtained as
\begin{equation}
     \tilde{R}_\Omega^k(q) = \frac{\sum\limits_{x \in P_q} \sigma_{\tau_1} (k  - 1 - \sum\limits_{\substack{z \in \Omega\\ z \neq x}} \sigma_{\tau_2}(s_{qz} - s_{qx}) )}{|P_q|},
\label{equ:smooth_recall}
\end{equation}
which is differentiable and can be used for training with back-propagation. The two sigmoids have different function domains and, therefore, different temperatures (see Figure~\ref{fig:sigmoids}). The minimized single-query loss in a mini-batch $B$, with size $M=|B|$, and query $q\in B$ is given by 
\begin{equation}
L^k(q) = 1- \tilde{R}_{B\setminus q}^k(q).
\end{equation}
while incorporation of multiple values of $k$ is performed in the loss given by 
\begin{equation}
L^K(q) = \frac{1}{|K|}\sum_{k \in K} L^k(q).
\label{equ:loss}
\end{equation}
Figure~\ref{fig:deriv} shows the impact of using single or multiple values for $k$. 

All examples in the mini-batch are used as queries and the average loss over all queries is minimized during the training.
The proposed loss is referred to as \emph{Recall@k Surrogate loss}, or RS@k loss for brevity.

To allow for 0 loss when $k$ is smaller than the number of positives (note that exact recall@k is less than 1 by definition), we slightly modify \equ{smooth_recall} during the training. Instead of dividing by $|P_q|$, we divide by $\min(k, |P_q|)$, and, consequently, we clip values larger than $k$ in the numerator to avoid negative loss values.

\subsection{Similarity mixup (SiMix)} Given original batch $B$, virtual batch $\hat{B}$ is created by mixing all pairs of positive examples in the original batch.
Embeddings of examples $x\in B$ and $z \in B$ are used to generate mixed embedding
\begin{equation}
    \vv_{xz_{\alpha}} = \alpha \vx + (1-\alpha) \vz \quad | \quad \alpha \sim U(0,1),
\end{equation}
for a virtual example that is denoted by $xz_{\alpha} \in \hat{B}$.
The similarity of an original example $w\in B$ to the virtual example $xz_{\alpha} \in \hat{B}$ is given by
\begin{equation}
s(w,xz_{\alpha}) = \vw^\top \vv_{xz_{\alpha}} = \alpha s_{wx} + (1-\alpha) s_{wz}, 
\label{equ:simix1} 
\end{equation}
where the original and virtual examples can be the query and database examples, respectively, or vice versa. 
In case both examples are virtual, \eg $xz_{\alpha_1} \in \hat{B}$ used as a query and $yw_{\alpha_2} \in \hat{B}$ as a part of the database, then their similarity is given by
\begin{align}
s(xz_{\alpha_1},yw_{\alpha_2}) &= \vv_{xz_{\alpha_1}}^\top \vv_{yw_{\alpha_2}} \nonumber\\ 
        &= \alpha_1 \alpha_2 s_{xy} + (1-\alpha_1) (1- \alpha_2) s_{zw} \nonumber\\
        &+ \alpha_1 (1-\alpha_2) s_{xw}+(1-\alpha_1) \alpha_2 s_{zy}.
\label{equ:simix2} 
\end{align}
The pairwise similarities that appear on the right-hand side of the previous formulas, \eg $s_{wx}$ and $s_{wz}$ in \equ{simix1}, are computed from the embeddings of the original, non-virtual examples and are also required for the computation of the RS@k without any virtual examples. Therefore, the mini-batch is expanded to $B \cup \hat{B}$ by adding virtual examples without the need for explicit construction of the corresponding embeddings or computation of the similarity via dot product; simple mixing of the corresponding pairwise scalar similarities is enough. SiMix reduces to mixing pairwise similarities due to the lack of re-normalization of the mixed embeddings, which is different to existing practice in prior work~\cite{vpa+21,gk+20,gkk+21,ksp+20} and brings training efficiency benefits.

\begin{figure}[t]
\begin{center}
\hspace{-5pt}
\includegraphics[width=0.48\textwidth]{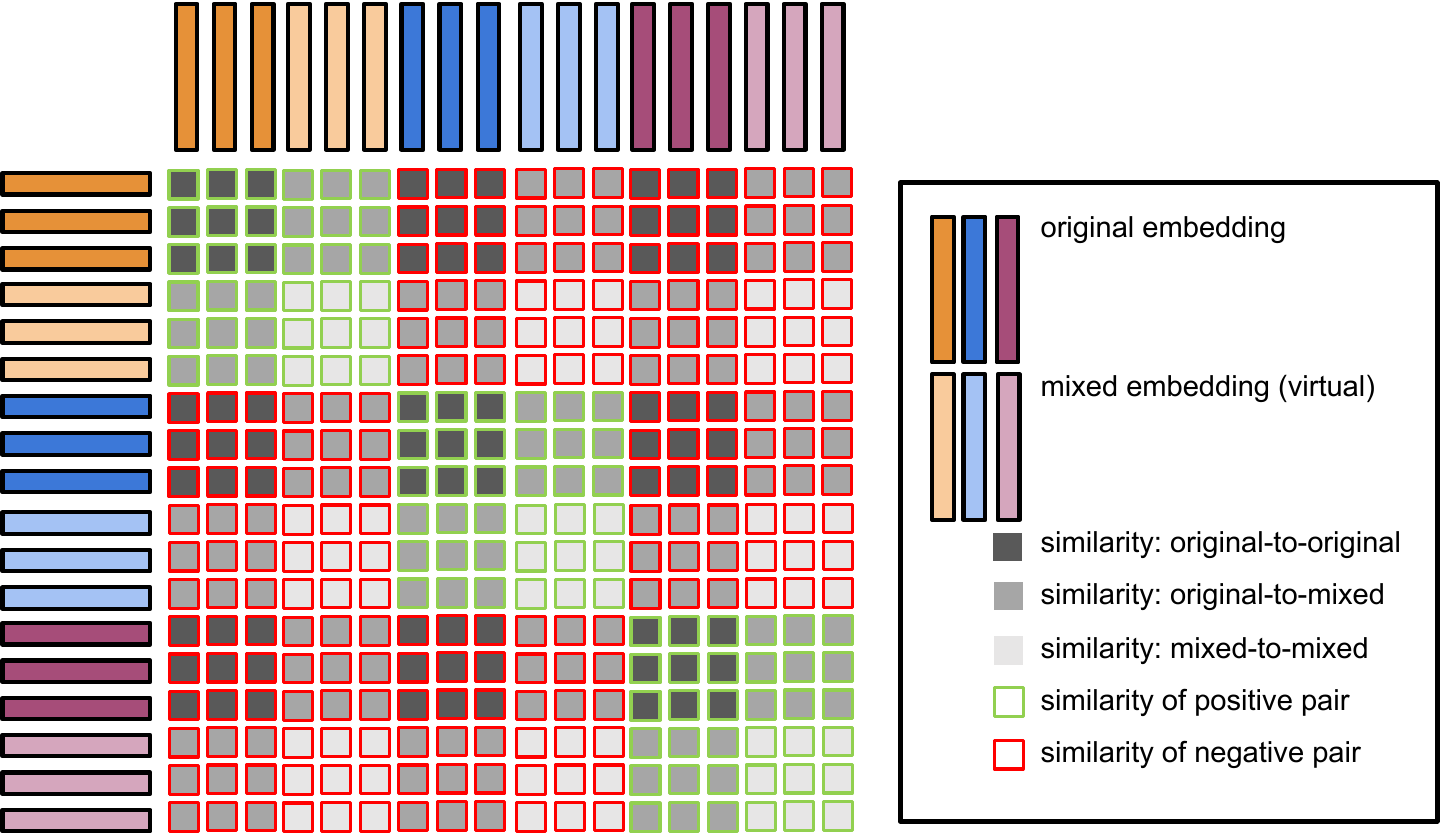}
\hspace{-15pt}
\caption{SimMix on a toy example with 3 classes (organge, blue, purple). The batch is comprised original and virtual examples. Three kinds of pairwise similarities are computed using direct dot product similarity, \equ{simix1}, or \equ{simix2}. The RS@k is applied independently per row of the similarity matrix  according to the positive/negative labels. Mixed embeddings are illustrated but are never expliticly created; all similarities are estimated directly from the original embeddings.
\label{fig:simmix}
}
\end{center}
\end{figure}

Virtual examples are created only between examples of the same classes and are labeled according to the class of the original examples that are mixed. Virtual examples are used both as queries and as database examples, while mixing is applied to all pairs of positive examples inside a mini-batch. 
Figure~\ref{fig:simmix} depicts an illustration showing the construction of the virtual samples.

\subsection{Large batch size}
Since recall@k is a non-decomposable objective function, which needs to be computed on the entire training dataset. To better reflect this objective function during training with the proposed surrogate loss, use of large batch sizes is required for the training. 
Additionally, the larger the batch the high the chance for hard negative examples to exist within the batch. 
Therefore, such a choice dispenses with the need for hard-negative mining which is common in DML~\cite{wms+17,see+15}.
Nevertheless, with limited number of GPUs, hardware constraints set restrictions to the batch size. 
We follow the multistage back-propagation proposed by Revaud~\etal~\cite{rar+19}. 
A forward pass is performed to obtain all embeddings, while intermediate tensors are discarded from memory/computational graph. 
Then, the loss is computed, and so are the gradients \wrt the embeddings. 
Finally, each of the embeddings is recomputed separately in a second forward pass, this time allowing the propagation of the gradients. 
This implementation detail is outlined in Algorithm~\ref{alg:largebatch} and is applicable beyond the proposed RS@k loss or the mean Average Precision loss~\cite{rar+19}.
The batch-size impact for the proposed RS@k loss function is validated in the experimental section.

\begin{algorithm}
\caption{Training with a very large batch.\label{alg:largebatch}}
\input{algo_largebatch}
\end{algorithm}

\begin{algorithm}
\caption{Training with RS@k and SiMix.\label{alg:main}}
\input{algo}
\end{algorithm}

\hspace{5pt}
\subsection{Pre-training method}
\label{sec:pretraining_method}
We briefly present the different pre-training methods that we use as model initialization for performing DML. The pre-training methods explored in this paper span supervised learning~\cite{dbk+21}, weakly supervised~\cite{sga+22}, self-supervised~\cite{odm+23} and text-supervised~\cite{rkh+21,rdk+23}. The pre-trained models of those methods are open source and widely used for a variety of tasks.
We consider the following methods. 

\paragraph{ImageNet-$21$k} ~\cite{dbk+21}. This variant corresponds to supervised training with cross-entropy loss on ImageNet-$21$k, which is a super-set of ILSVRC-$2012$ ImageNet dataset~\cite{dsl+09}. The dataset contains $21$k classes and $14$M image in total. 

\paragraph{SWAG}~\cite{sga+22} performs weakly-supervised training with multi-class cross-entropy loss on a large scale dataset of images and the corresponding hashtags. The dataset comprises  $3.6$ billion images and hashtags from Instagram, with a total of $27$K canonical hashtags. The data is pre-processed to balance the hashtags frequency, ensuring less frequent tags are not overshadowed by the more common ones. 

\paragraph{DINOv2}~\cite{odm+23} performs self-supervised learning on a curated dataset of $142$ million images, which is refined through an automatic pipeline that ensures diversity and quality, avoiding biases of uncurated datasets. The training objective is a combination of several loss functions, including image-level DINO~\cite{ctm+21} and patch-level iBOT losses~\cite{zww+21} with SwAV's centering~\cite{cmm+20} for robust feature learning across different image patches. 

\paragraph{CLIP}~\cite{rkh+21}. is a text-supervised learning approach of a visual and textual encoder models to associate images with corresponding textual information. The training is performed on large scale datasets from the web, such as the LAION-400M and LAION-2B datasets~\cite{iww+21,cbw+23}, with $400$ million and $2$ billion image alt-text pairs respectively. CLIP uses a contrastive loss function that maximizes the similarity between co-occurring image-text pairs while minimizing it for mismatched pairs.

\paragraph{DiHT}~\cite{rdk+23} is another text-supervised approach like CLIP that focuses on enhancing vision-language models through three key improvements over CLIP. The training set is based on LAION-$2$B and is filtered down to $438$ million samples with a new dataset filtring method. In addition to the filtered data, a concept distillation method is used to leverage pre-trained uni-modal (vision only) representations to enhance multi-modal training. Additionally, the contrastive learning objective of CLIP is modified to incorporate an importance-sampling strategy for hard negatives.

\hspace{5pt}
\subsection{Overview} An overview of the training process with the proposed loss and SiMix is given in Algorithm \ref{alg:main}. In case SiMix is not used, then lines \ref{lin:vbatch}, \ref{lin:vsim1}, \ref{lin:vsim2} and \ref{lin:expand} are skipped. It is assumed that each image in training is labeled to a class. Mini-batches of size $M$ are generated by randomly sampling $m$ images per class out of $\nicefrac{M}{m}$ randomly sampled classes.
\hspace{5pt}

%% file: fig_sigmoid.tex
\begin{tikzpicture}[declare function={sigma(\x)=1/(1+exp(-\x));sigmap(\x)=sigma(\x)*(1-sigma(\x));}]
\begin{axis}[%
	width=0.57\linewidth,
	height=0.5\linewidth,
	xlabel={$u = k -1 - r_\Omega(q,x)$},
	title={$g(u)= \sigma_{\tau_1}(u), \tau_1=1$},
	legend cell align={left},
	legend pos=north west,
    legend style={cells={anchor=east}, font =\tiny, fill opacity=0.8, row sep=-2.5pt},
    xmin = -8,
    xmax = 8,
    domain=-8:8
]
\addplot[blue,mark=none,samples=500]   (x,{sigma(x/1.0)});\addlegendentry{$g(u)$}
\addplot[red,mark=none,samples=500]   (x,{sigmap(x/1.0)});\addlegendentry{$\frac{dg(u)}{du}$}
\end{axis}
\end{tikzpicture}
\hspace{-5pt}
{
\begin{tikzpicture}[declare function={sigma(\x)=1/(1+exp(-\x));sigmap(\x)=sigma(\x)*(1-sigma(\x));}]
\begin{axis}[%
	width=0.57\linewidth,
	height=0.5\linewidth,
	xlabel={$u= s_{qz}-s_{qx}$ },
	title={$g(u)= \sigma_{\tau_2}(u), \tau_2 =0.01$},
	legend cell align={left},
	legend pos=north west,
    legend style={cells={anchor=east}, font =\small, fill opacity=0.8, row sep=-2.5pt},
    xmin = -1,
    xmax = 1,
    domain=-1:1
]
\addplot[blue,mark=none,samples=500]   (x,{sigma(x/0.01)});
\addplot[red,mark=none,samples=500]   (x,{sigmap(x/0.01)});
\end{axis}
\end{tikzpicture}
}

%% file: fig_deriv.tex
\begin{tikzpicture}[declare function={sigma(\x)=1/(1+exp(-\x));sigmap(\x)=sigma(\x)*(1-sigma(\x));}]
\begin{axis}[%
	width=0.99\linewidth,
	height=0.5\linewidth,
	xlabel={$r$},
	ylabel={$\frac{d\sigma_{\tau_1}(u)}{du},~u=k-r$},
	legend cell align={left},
	legend pos=north east,
    legend style={cells={anchor=east}, font =\tiny, fill opacity=0.8, row sep=-2.5pt},
    xtick = {1,3,5,7,9},
    xmin = 1,
    xmax = 10,
    domain=0:30
]
\addplot[blue,mark=none,samples=500]   (x,{sigmap(x-1)});\addlegendentry{$k=1$}
\addplot[red,mark=none,samples=500]   (x,{sigmap(x-2)});\addlegendentry{$k=2$}
\addplot[green,mark=none,samples=500]   (x,{sigmap(x-4)});\addlegendentry{$k=4$}
\addplot[black,mark=none,samples=500]   (x,{sigmap(x-1)+sigmap(x-2)+sigmap(x-4)});\addlegendentry{$K=\{1,2,4\}$};
\end{axis}
\end{tikzpicture}

%% file: algo_largebatch.tex
\footnotesize
\algrenewcommand\algorithmicindent{0.5em}%
\begin{algorithmic}[1]
\Procedure{Train-large-batch}{$X$, $Y$, $f$, $\ell$, \text{optimizer}}
\color{gray}
\State $X: $ batch of $M$ training images
\State $Y: $ batch class labels
\State $f: $ model
\State $\ell: $ loss function 
\color{black}
\State
\State optimizer.zero\_grad()
\State with nograd(): $\mathbf{X} \gets f(X)$ \comment{extract batch embeddings}
\State $\mathbf{X.\text{detach()}}$ \comment{remove from the computational graph}
\State $l=\ell(\mathbf{X}, Y)$ \comment{compute the total loss}
\State $l.\text{backward()}$ \comment{backward pass - up to the embeddings}
\State Xgrad $ = \mathbf{X}.\text{grad}$ \comment{maintain the loss gradient \wrt embeddings}
\State del $\mathbf{X}$ \comment{remove embeddings from memory}
\For{$i \in 1 \ldots M$} \comment{one image at a time}
\State $\vx \gets f(X[i])$  \comment{re-extract image embedding}
\State $\vx.\text{backward(Xgrad[i])}$ \comment{continue the backward pass}
\EndFor
\State optimizer.step() \comment{model update}
\EndProcedure
\end{algorithmic}

%% file: algo.tex
\footnotesize
\algrenewcommand\algorithmicindent{0.5em}%
\begin{algorithmic}[1]
\Procedure{Train-RS@k}{$X$, $Y$, $M$, $m$}
\color{gray}
\State $X: $ training images
\State $Y: $ class labels
\State $M: $ mini-batch size
\State $m: $ number of images per class in mini-batch
\color{black}
\State
\State $\theta \gets$ pre-training initialization \comment{CLIP, DiHT, SWAG, DINOv2, ImageNet-21k}
\For{$\text{iteration} \in [1,\ldots, \text{number-of-iterations}]$}
\State $loss \gets 0$ \comment{set batch loss to zero}
\State $B \gets$ \Call{Batch-sampler}{$X$, $Y$, $M$, $m$} 
\State $\hat{B} \gets$ \Call{Virtual-batch}{$B$} \comment{enumerate virtual examples} \label{lin:vbatch}
\State \algorithmicfor { $(x,z) \in B \times B$} \algorithmicdo { compute $s(x,z)$ }  \comment{use $\vx^\top \vz$} 
\State \algorithmicfor { $(x,z) \in B \times \hat{B}$} \algorithmicdo { compute $s(x,z)$ }  \comment{use \equ{simix1}} \label{lin:vsim1}
\State \algorithmicfor { $(x,z) \in \hat{B} \times \hat{B}$} \algorithmicdo { compute $s(x,z)$ }  \comment{use \equ{simix2}} \label{lin:vsim2}
\State $B \gets B \cup \hat{B}$ \comment {expand batch with virtual examples} \label{lin:expand}
\For{$q \in B$} \comment {use each image in the batch as query}
\State $loss \gets loss + L^K(q)$ \comment {Recall@k loss \equ{loss}}
\EndFor
\State $\theta \gets $ \Call{Minimize}{$\frac{loss}{|B|}$} \comment {SGD update}
\EndFor
\EndProcedure
\end{algorithmic}

%% file: 3_experiments.tex
\section{Experiments}
\label{sec:experiments}

\subsection{Datasets}
\label{sec:datasets}

\begin{table}
\centering
\small
\begin{tabular}{l|r|r|r}
    \hline
    \textbf{Dataset} & \textbf{\#Images} &  \textbf{\#Classes} & \textbf{\#Avg}\\
    \hline\hline
    iNaturalist Train~\cite{vms+18} & $325,846$ & $5,690$ & $57.3$ \\
    iNaturalist Test~\cite{vms+18} & $136,093$ & $2,452 $ & $55.5$ \\
    VehicleID Train~\cite{ltw+16} & $110,178$ & $13,134$ & $8.4$ \\
    VehicleID Test~\cite{ltw+16} & $40,365 $ & $4,800$ & $8.4$ \\
    SOP Train~\cite{ohb16} & $59,551 $ &  $11,318$ & $5.3$ \\
    SOP Test~\cite{ohb16} & $60,502$ &  $11,316$ & $5.3$ \\
    Cars196 Train~\cite{ksd+13} & $8,054$ & $98$ & $82.1$ \\
    Cars196 Test~\cite{ksd+13} & $8,131$ & $98$ & $82.9$ \\
    \hline
    $\mathcal{R}$Oxford~\cite{rit+18} & $4,993$ & $11$ & n/a \\
    $\mathcal{R}$Paris~\cite{rit+18} & $6,322$ & $11$ & n/a \\
    GLDv1~\cite{nas+17} & $1,060,709$ & $12,894$ & $82.3$ \\
    \hline
\end{tabular}
\caption{Dataset composition for training and evaluation.}
\label{tab:datasets}
\end{table}

The training and evaluation is performed on four widely used deep metric learning benchmarks, namely iNaturalist~\cite{vms+18}, PKU VehicleID~\cite{ltw+16}, Stanford Online Products~\cite{ohb16} (SOP), and Stanford Cars~\cite{ksd+13} (Cars196). Recall at top $k$ retrieved images,  denoted by r@k, is one of the standard evaluation metrics in these benchmarks.  Metric r@k is 1 if at least one positive image appears in the top $k$ list, otherwise 0. The metric is averaged across all queries. Note that this is different from the standard definition of recall in \equ{recall}. 

iNaturalist~\cite{vms+18} is firstly used by Brown~\etal~\cite{bxk+20}, whose setup we follow: $5,690$ classes for training and $2,452$ classes for testing. For VehicleID, according to the standard setup~\cite{ltw+16}, $13,134$ classes are used for training, and the evaluation is conducted on the predefined small ($800$ classes), medium ($1600$ classes) and large ($2400$ classes) test sets. For SOP~\cite{ohb16} and Cars196~\cite{ksd+13}, the standard experimental setup of Song~\etal~\cite{sxj+15} is followed. The first half of the classes are used for training and the rest for testing, resulting in $11,318$ classes for SOP and $98$ for Cars196.

The method is evaluated for instance-level search on Revisited Oxford ($\mathcal{R}$Oxford) and Paris ($\mathcal{R}$Paris) benchmark~\cite{rit+18}, where the evaluation metric is mean Average Precision (mAP). The training uses the Google Landmarks dataset (GLDv1)~\cite{nas+17} to perform a comparison with the work of Revaud~\etal~\cite{rar+19} and their AP loss. The validation is performed according to the work of Tolias~\etal~\cite{tjc20}. 

The number of examples, classes, and average number of examples per class can be found in Table \ref{tab:datasets}. Note that these datasets are diverse in the number of training examples, the number of classes, and the number of examples per class, ranging from class balanced~\cite{ksd+13} to long-tailed~\cite{vms+18}.

\subsection{Implementation Details}
\label{sec:implementation_details}

We experiment with CNN and ViT based vision encoders. The implementation details differ based on the vision encoder used for an experiment. The implementation details are identical across the metric learning benchmarks but differ for  $\mathcal{R}$Oxford/$\mathcal{R}$Paris to follow and compare to prior work~\cite{rar+19}. These differences are clarified when applicable.

\paragraph{Architectures.} For CNN based experiments, an ImageNet~\cite{dsl+09} pre-trained ResNet-50~\cite{hzr+16} is used as the backbone for deep image embeddings. Building on the standard implementation of~\cite{rms+20}, the BatchNorm parameters are kept frozen during the training. After the convolutional layers, Generalized mean pooling~\cite{rtc19} and layer normalization~\cite{bkh+16} are used, similar to~\cite{tdt20}. The last layer of the model is a $d$ dimensional fully connected (FC) layer with $L_{2}$ normalization. In the case of $\mathcal{R}$Oxford/$\mathcal{R}$Paris, ResNet-101~\cite{hzr+16} is used, layer normalization is not added, while the FC layer is initialized with the result of whitening~\cite{rtc19}.
For vision transformers~\cite{dbk+21} (ViT) based setup, we experiment with different initialization based on the encoder pre-training. For model implementation and initialization, we rely on PyTorch hub and timm libraries~\cite{rw2019timm}. The details of explored initialization for ViT encoders are presented in Section ~\ref{sec:pretraining_method}. For CLIP and DiHT based models, we only use the vision encoders, while the text encoders are discarded. We use the representations corresponding to the [CLS] token and use a linear projection layer to match to the desired dimension.

\paragraph{Training details.} For ResNet architectures, Adam optimizer~\cite{kb15} is used and for ViTs, AdamW~\cite{lh+19} is used. This paper follows the standard class-balanced-sampling~\cite{mbl20,bxk+20,tdt20} with $4$ samples per class for all the datasets, while classes with less than $4$ samples are not used for training. Unless stated otherwise, the batch size for training is set $4,000$ for all datasets but Cars196 where it is equal to $4*\text{{\em\#classes}}=392$. 
When training on GLDv1 and testing on $\mathcal{R}$Oxford/$\mathcal{R}$Paris, the batch size is set to $4096$~\cite{rar+19}, and training is performed for 500 batches, while other training hyper-parameters are set as in the work and GitHub implementation of Radenovic \etal~\cite{rtc19}. Note that the hyper-parameters for each dataset are released with our implementation.

\paragraph{RS@k hyper-parameters.} The proposed Recall@k Surrogate (RS@k) loss \equ{smooth_recall} contains three hyper-parameters: sigmoid temperature $\tau_{2}$ - applied on similarity differences, sigmoid temperature $\tau_{1}$ - applied on ranks and the set of values for $k$ for which the loss is computed. Both sigmoid temperatures are kept fixed across all the experiments as $\tau_{2}=0.01$ (same as~\cite{bxk+20}) and $\tau_{1}=1$. The values of $k$ are kept fixed as $k=\{1,2,4,8,16\}$ without SiMix and $k=\{1,2,4,8,12,16,20,24,28,32\}$ with SiMix. For GLDv1~\cite{nas+17}, this is $k=\{1,2,4\}$, and $k=\{1,2,4,8\}$, respectively. The values of $k$ and the sigmoid temperature $\tau_{1}$ are investigated in our experiments, where it is observed that the method is not very sensitive to these hyper-parameters. 


\paragraph{Validation protocol - hyper-parameter tuning.}
In the standard split, the metric learning benchmarks~\cite{ksd+13,ohb16,ltw+16,vms+18} do not contain an explicit validation set; as a result, image retrieval methods often tune the hyper-parameters on the test set, leading to the issue of training with test set feedback. This issue has been studied in~\cite{mbl20}, which proposes to train different methods with identical hyper-parameters. The setup of~\cite{mbl20} is not directly usable for experiments with the RS@k loss, as large batch sizes are crucial to estimate recall@k accurately. Furthermore, their setup does not allow mixup.

As a remedy, following the setup of ProxyNCA++~\cite{tdt20}, the training set is split into training and validation by using the first half of the classes for training and the other half for validation. With this split, a randomized grid search determines the learning rate, decay steps, decay size and the total number of epochs. Once the hyper-parameters are fixed, training is conducted once on the entire training set and evaluated on the test set. 
Note that not all methods we compare with follow this principled protocol for hyper-parameter tuning. 
It appears that wrong practice is commonly followed in the literature~\cite{mbl20}, making comparisons across papers unfair.

\paragraph{Experimental details.}
The methods in the literature use different embedding sizes, $d$, therefore, the models for the RS@k loss are trained with two embedding sizes of $d=128$ and $d=512$ for metric learning benchmarks \cite{vms+18,ltw+16,ohb16,ksd+13}, and $d=2048$ for $\mathcal{R}$Oxford/$\mathcal{R}$Paris \cite{rit+18}, to allow a fair comparison. For ViT based experiments, we always use $d=512$ for simplicity.

Unless otherwise stated, the results of the competing methods are taken from the original papers. Methods marked with a $\dag$ were trained by the authors of this paper, using the same implementation as used for the RS@k loss. The results on metric learning benchmarks~\cite{ksd+13,ohb16,ltw+16,vms+18} are compared with the methods that use either ResNet-50~\cite{hzr+16} or Inception network~\cite{slj+15}. ResNet-50~\cite{hzr+16} is represented as $R_{50}^{d}$ in the tables and the standard Inception network~\cite{slj+15} as $I_{1}^{d}$, the Inception network with BatchNorm as  $I_{3}^{d}$ (same as ~\cite{tdt20}). Here $d$ is the embedding size. On all the datasets, the performance of the baseline, Smooth-AP (SAP)~\cite{bxk+20}, is also reported with Generalized mean pooling~\cite{rtc19} and layer normalization~\cite{bkh+16}, shown as SAP\textsuperscript{\dag} (+Gem +LN). This is to eliminate any performance boost in the comparisons that were caused by the architecture. Note that unless otherwise stated in our experiments, the batch size for SAP is set as $384$, the same as the original implementation~\cite{bxk+20}. 
Further, we demonstrate the performance of SAP and the proposed RS@k on ViT-B architectures initialized with ImageNet-21k pre-training. 
The variant of ViT-B that uses a patch size of $32\times32$ is denoted by ViT-B/32 and the one that uses a patch size of $16\times16$ by ViT-B/16.
Similarly, we use ViT-L/14 and ViT-L/16. For experiments with different initializations, we use ViT-B/32, ViT-B/16 and ViT-L/16. For some initialization such as SWAG, DiHT, CLIP and DiNOv2, we use ViT-L/14, that is, a patch-size of $14\times14$.

\subsection{Results}
\label{sec:evaluation}

\begin{figure*}
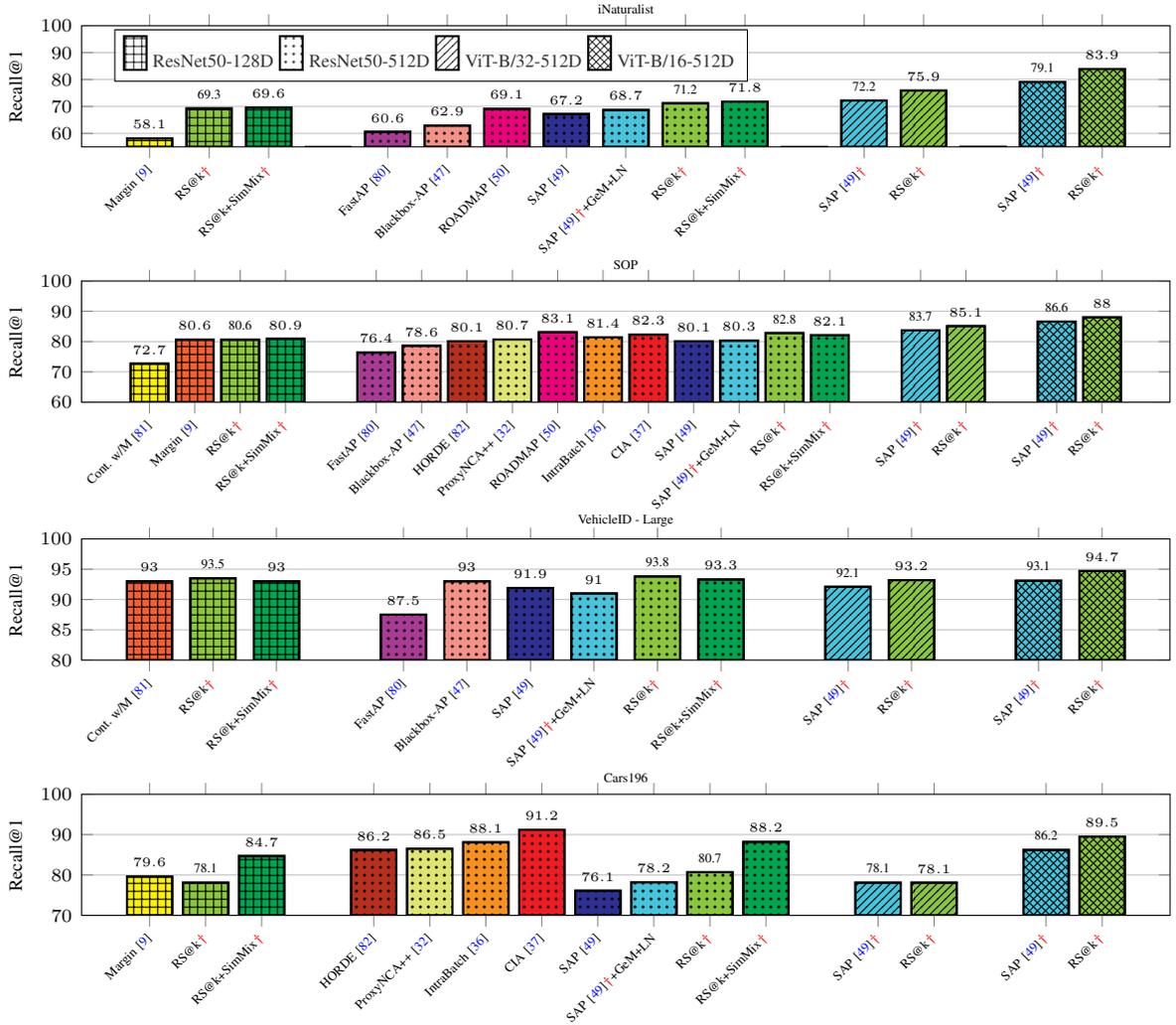

\vspace{-20pt}
\centering
\include{results_barplot}
\vspace{-25pt}
\caption{\textbf{Performance comparison with other methods.} Recall@1 on iNaturalist, SOP, VehicleID and Cars196. Methods marked with $\dag$ were trained using the same pipeline by the authors of this paper.
\label{fig:barplot}}
\end{figure*}

\begin{figure*}
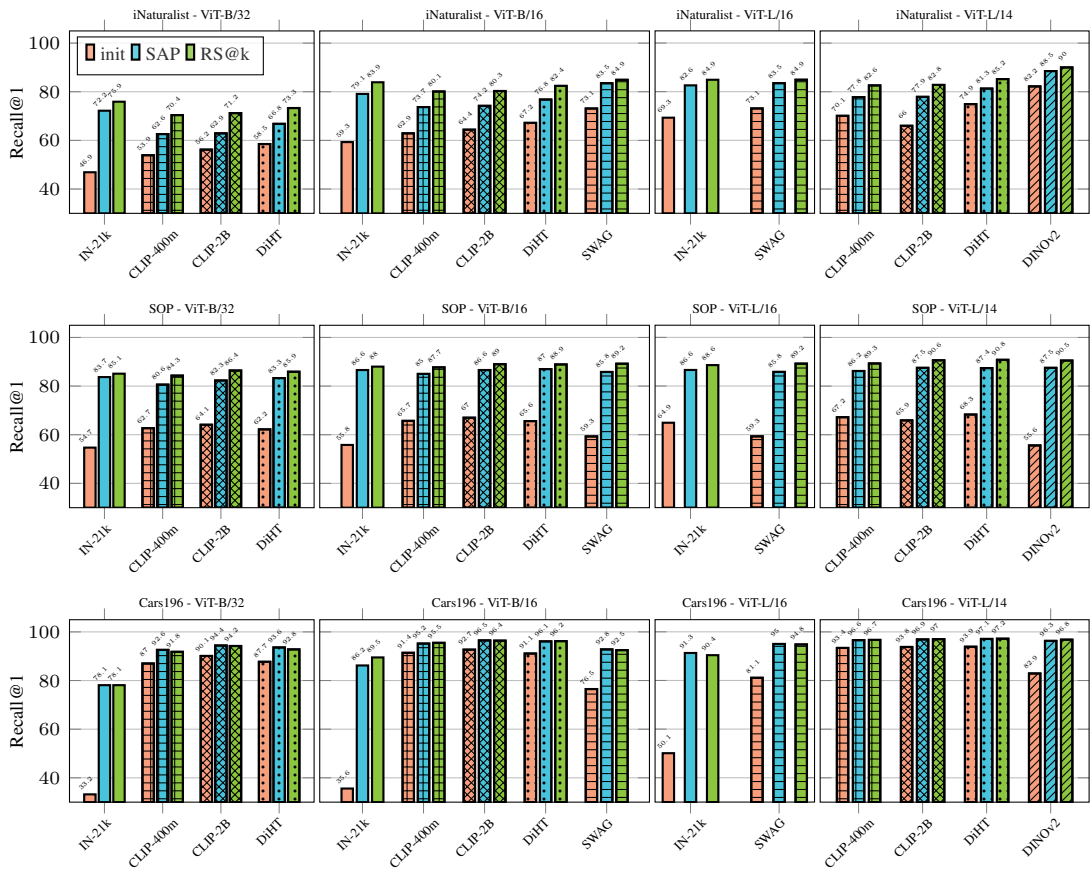

\vspace{-20pt}
\centering
\include{results_barplot_init}
\vspace{-30pt}
\caption{\textbf{The impact of model initialization}. Performance comparison between SAP and RS@k for a variety of backbones and methods used as pre-training. The performance at the moment of initialization is shown for reference. Recall@1 on iNaturalist, SOP, and Cars196. 
\label{fig:barplot_init}}
\end{figure*}

\paragraph{Comparison to SoTA.}
We present the performance comparison to existing methods with a variety of backbones in Figure~\ref{fig:barplot}. 
A major focus is on performing a direct comparison with SAP, which is a very similar method and a main competitor. 
We perform a large set of experiment with SAP by ourselves under fair conditions. 
The comparison shows that RS@k is consistently surpassing SAP across backbones, dimensionalities and datasets. 
The use of SimMix ranges from small improvements and small performance drop to a very large that happens on Cars16. 
Due to the small number of classes in this dataset, the batch size is relatively small (392). 
Therefore, the virtual batch size increase performed by SimMix brings it to a much larger scale which makes the loss effective. 
We conclude that given the tiny extra cost of SimMix it is better to use it than to skip it, especially if the original batch size is relatively small, \eg for small datasets or due to hardware restrictions. 
Compared to all other approaches, RS@k appears to achieve the top performance with few exceptions. 
ROADMAP is surpassing RS@k by 0.3 on SOP but not on iNaturalist, while CIA achieves the best result on Cars196, but not on SOP.
The cross-image attention contribution of CIA~\cite{kpm+23} is orthogonal to ours contributions in this work, while the same holds for the piece-wise asymmetric approximation of the step function used in ROADMAP.
Note that other than this work, to our knowledge only ProxyNCA++~\cite{tdt20} uses a principled hyper-parameter tuning protocol, while for the rest of the methods, it is unknown how were the hyper-parameters tuned.
A more detailed comparison is presented in Table~\ref{tab:metriclearning} at the end of this manuscript including evaluation of recall at windows of size larger than 1.

\paragraph{The impact of model initialization.}
Figure~\ref{fig:barplot_init} presents an extensive comparison of different model initializations, relying on different pre-training approaches for a variety of ViT backbones. 
The results include both our RS@k and SAP, demonstrating once more the superiority of RS@k for the majority of the cases.
Some observations follow.
IN-21k provides a lower starting point and despite the large improvements after training and catching-up or surpassing other variants in some of the cases, in most of the cases it does not result in a top-ranked model. Therefore, recent large-scale pre-training methods that are not fully supervised as effective initializations for DML.
%
%
The larger CLIP training set (2B) always results in better starting and final point compared to the the smaller one (400m).
DiHT shows good improvements over CLIP-2B on iNaturalist only, but shows improvements over CLIP-400m on all datasets. Note that the final training set of DiHT is roughly 400m.
DINOv2 is the best performing on iNaturalist by a large margin, but on par with other variants on the other datasets. 
%
%
A more detailed comparison is presented in Table~\ref{tab:metriclearning_initialization} at the end of this manuscript including evaluation of recall at windows of size larger than 1.
Using a strong backbone and good initialization, we are able to achieve 97.6 Recall@16 on iNaturalist, 97.7 Recall@10 on SOP, and 99.3 Recall@8 Cars196, which signifies that under these settings those benchmarks are nearly solved.


\paragraph{Results on instance-level search - $\mathcal{R}$Oxford/$\mathcal{R}$Paris.} Table~\ref{tab:rop} summarizes a comparison with AP-based losses in the literature on $\mathcal{R}$Oxford/$\mathcal{R}$Paris with and without distractor images. The comparison is performed with GLDv1 as a training set whose performance is reported for the work of Revaud \etal~\cite{rar+19} in their GitHub page.
During the training performed by us, images are down-sampled to have a maximum resolution of $1024 \times1024$. 
The inference is performed with multi-resolution descriptors at three scales with up-sampling and down-sampling by a factor of $\sqrt{2}$.
Note that SAP is not evaluated on these datasets in the original work and this experiment is performed by us, which outperforms the previously used AP loss~\cite{hls18}. RS@k, with or without the SiMix, increases the performance by a small margin.

\begin{table*}[t!]
\hspace{5pt}
\begin{center}
\resizebox{0.99\textwidth}{!}{\input{tab_rop}}
\hspace{5pt}
\caption{Performance comparison (mAP\%) on $\mathcal{R}$Oxford and $\mathcal{R}$Paris with 1m distractor images ($\mathcal{R}$1m). Mean performance is reported across all setups or the large-scale setups only. $\ast$ denotes that the FC layer is not part of the training but is added afterwards to implement whitening. Batch size is 4096 for all methods; SiMix virtually increases it to 10240. ResNet101 is used as a backbone for all methods.
\label{tab:rop}}
\end{center}
\end{table*}



\begin{figure}[b]
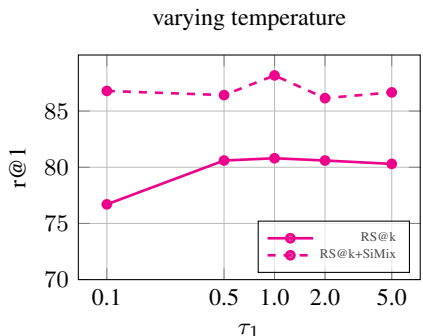

\vspace{-10pt}
\centering
\include{fig_temp}
\vspace{-30pt}
\caption{The effect of sigmoid temperature $\tau_{1}$ applied on ranks. Results are shown on Cars196~\cite{ksd+13}.}
\label{fig:temp}
\end{figure}

\paragraph{Impact of the sigmoid temperature $\tau_{1}$ - applied on ranks.}
The effect of the sigmoid temperature $\tau_{1}$ is summarized in Figure \ref{fig:temp}. For both setups of with and without SiMix, $\tau_{1}=1.0$ gives best results while higher and lower values lead to a decline.
A very small value corresponds to a very sharp sigmoid and zero gradients over a large range of values, which results in the lowest performance.
Note that SimMix performs well for the whole range of values shown.

\begin{figure}[b]
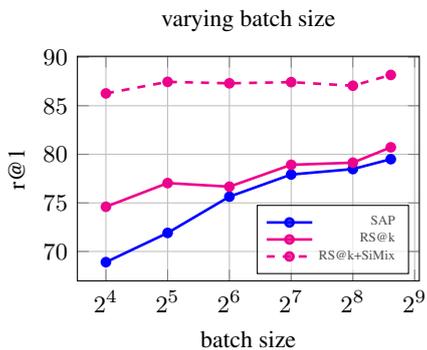

\vspace{-10pt}
\centering
\include{fig_bs}
\vspace{-30pt}
\caption{The effect of batch size. Results are shown on Cars196~\cite{ksd+13}.}
\label{fig:bs}
\end{figure}

\paragraph{Impact of the batch size.}
The effect of the varying batch size is shown in Figure \ref{fig:bs}. It demonstrates that large batch size leads to better results. A significant performance boost is observed with the use of SiMix, especially in the small batch size regime, which comes at a small extra computation. 
SimMix, due to the virtual batch size increase, has a large performance boost even in the small batch regime.
Note that we show results for Cars196 where SimMix has the best impact among other datasets.

\paragraph{Impact of {\em k}.} The study for the set of values of {\em k} used for RS@k loss can be found in Table \ref{tab:Cars196_ablation}. The results RS@$\{1\}$, RS@$\{1,2\}$, RS@$\{1,2,4\}$ and RS@$\{1,2,4,8\}$ suggest that adding larger values of {\em k} leads to a small decline in the performance. However, RS@$\{1,2,4,8,16\}$ gives on average the same results as RS@$\{1\}$, with higher performance on larger {\em k} values. Comparing entries RS@$\{4,8,16\}$ and RS@$\{1,2,4,8,16\}$ suggests that the use of small values, such as $k=1$ or $k=2$, is important as the performance drops when these values are removed. Further removing $k=4$ (RS@$\{8,16\}$) does not change the performance. However, removing $k=8$ (RS@$\{16\}$) leads to a significant decline in the performance.

\begin{table}[t]
    \input{tab_Cars196_ab}
    \caption{Varying the set of values of {\em k}. Results on Cars196~\cite{ksd+13}. In all experiments, $\tau_{1}=1$ and $\tau_{2}=0.01$.}
    \label{tab:Cars196_ablation}
\end{table}

\begin{table*}[t]
\begin{center}
\footnotesize
\setlength{\tabcolsep}{1.5pt}
    \resizebox{0.99\textwidth}{!}{\input{tab_metriclearning}}
    \caption{\textbf{Performance comparison with other methods.} Among the convolutional backbones results, the best and second-best results are shown with \textbf{bold} and \underline{underline}, respectively. Methods marked with $\dag$ were trained using the same pipeline by the authors of this paper. Recall@$k$ is reported with the value of $k$ mentioned per column.}
    \label{tab:metriclearning}
\end{center}
\end{table*}

\begin{table*}[t]
\footnotesize
\begin{center}
    \resizebox{0.99\textwidth}{!}{\input{tab_metriclearning_initialization}}
    \caption{\textbf{The impact of model initialization}. Performance comparison between SAP and RS@k for a variety of backbones and methods used as pre-training. The performance at the moment of initialization is shown for reference. Best and second-best results per column are shown with \textbf{bold} and with \underline{underline}. Recall@$k$ is reported with the value of $k$ mentioned per column.}
    \label{tab:metriclearning_initialization}
\end{center}
\end{table*}


%% file: results_barplot.tex
\begin{tabular}{c}
\begin{tikzpicture}
    \begin{axis} [
        width  = \linewidth,
        height = 3.2cm,
        font=\footnotesize,
        symbolic x coords={Margin, RS@k0, RS@k+SimMix0, a, FastAP,BlackboxAP, ROADMAP, SAP, SAP1, RS@k1, RS@k+SimMix1,b,SAP2, RS@k2,c,SAP3, RS@k3},
        xtick={Margin, RS@k0, RS@k+SimMix0, FastAP,BlackboxAP, ROADMAP, SAP, SAP1, RS@k1, RS@k+SimMix1, SAP2, RS@k2,SAP3, RS@k3},
        xticklabels={Margin~\cite{wms+17}, RS@k{\color{red}\textbf{$\dagger$}}, RS@k+SimMix{\color{red}\textbf{$\dagger$}}, FastAP~\cite{chx+19},Blackbox-AP~\cite{rmp+20}, ROADMAP~\cite{ramzi2021robust}, SAP~\cite{bxk+20}, SAP~\cite{bxk+20}{\color{red}\textbf{$\dagger$}}{\tiny +GeM+LN}, RS@k{\color{red}\textbf{$\dagger$}}, RS@k+SimMix{\color{red}\textbf{$\dagger$}}, SAP~\cite{bxk+20}{\color{red}\textbf{$\dagger$}}, RS@k{\color{red}\textbf{$\dagger$}}, SAP~\cite{bxk+20}{\color{red}\textbf{$\dagger$}}, RS@k{\color{red}\textbf{$\dagger$}} },
        title = {iNaturalist},
        title style = {font=\tiny, yshift=-5pt},        
        x tick label style={rotate=45, align=right, text width=80pt, font=\tiny, xshift=-18pt, yshift=22pt},
        ylabel = {Recall@1},
        legend pos=north west,legend style={legend columns=-1, font=\footnotesize},
        ymin=55,ymax=100,
        bar width=0.6cm,                       
        ybar=-0.6cm,                           
        enlarge x limits={abs=0.9cm},
        nodes near coords,                     
        nodes near coords align={vertical},    
        every node near coord/.append style={font=\tiny,scale=1},        
        grid=none,
        ymajorgrids=true,xmajorticks=true,xminorticks=false,yminorgrids=true,
        legend image code/.code={\draw [#1] (0cm,-0.1cm) rectangle (0.3cm,0.35cm); },
]
        \addplot[style={black,fill=none,mark=none,pattern=grid}] coordinates {(a,0)}; 
        \addlegendentry{ResNet50-128D}
        \addplot[style={black,fill=none,mark=none,pattern=dots}] coordinates {(b,0)}; 
        \addlegendentry{ResNet50-512D}
        \addplot[style={black,fill=none,mark=none,pattern=north east lines}] coordinates {(c,0)}; 
        \addlegendentry{ViT-B/32-512D}
        \addplot[style={black,fill=none,mark=none,pattern=crosshatch}] coordinates {(c,0)}; 
        \addlegendentry{ViT-B/16-512D}
        %
        \addplot[style={black,fill=Yellow,mark=none,postaction={pattern=grid, pattern color=black}}] coordinates {(Margin, 58.1)};
        \addplot[style={black,fill=LimeGreen,mark=none,postaction={pattern=grid, pattern color=black}}, nodes near coords=69.3] coordinates {(RS@k0,69.3)};
        \addplot[style={black,fill=Green,mark=none,postaction={pattern=grid, pattern color=black}}] coordinates {(RS@k+SimMix0,69.6)};
        \addplot[style={black,fill=Mulberry,mark=none,postaction={pattern=dots, pattern color=black}}] coordinates {(FastAP, 60.6)};            
        \addplot[style={black,fill=Salmon,mark=none,postaction={pattern=dots, pattern color=black}}] coordinates {(BlackboxAP, 62.9)};
        \addplot[style={black,fill=RubineRed,mark=none,postaction={pattern=dots, pattern color=black}}] coordinates {(ROADMAP, 69.1)};
        \addplot[style={black,fill=Blue,mark=none,postaction={pattern=dots, pattern color=black}}] coordinates {(SAP,67.2)};
        \addplot[style={black,fill=SkyBlue,mark=none,postaction={pattern=dots, pattern color=black}}] coordinates {(SAP1,68.7)};
        \addplot[style={black,fill=LimeGreen,mark=none,postaction={pattern=dots, pattern color=black}}, nodes near coords=71.2] coordinates {(RS@k1,71.2)};
        \addplot[style={black,fill=Green,mark=none,postaction={pattern=dots, pattern color=black}}] coordinates {(RS@k+SimMix1,71.8)};        
        \addplot[style={black,fill=SkyBlue,mark=none,postaction={pattern=north east lines, pattern color=black}}, nodes near coords=72.2] coordinates {(SAP2,72.2)};
        \addplot[style={black,fill=LimeGreen,mark=none,postaction={pattern=north east lines, pattern color=black}}] coordinates {(RS@k2,75.9)};        
        \addplot[style={black,fill=SkyBlue,mark=none,postaction={pattern=crosshatch, pattern color=black}}, nodes near coords=79.1] coordinates {(SAP3,79.1)};
        \addplot[style={black,fill=LimeGreen,mark=none,postaction={pattern=crosshatch, pattern color=black}}] coordinates {(RS@k3,83.9)};        
    \end{axis}
\end{tikzpicture}
\\[-35pt]
\begin{tikzpicture}
    \begin{axis} [
        width  = \linewidth,
        height = 3.2cm,
        font=\footnotesize,
        symbolic x coords={Margin, ContwM, RS@k0, RS@k+SimMix0, a, FastAP,BlackboxAP,  HORDE,ProxyNCAplus, ROADMAP, IntraBatch, CIA, SAP, SAP1, RS@k1, RS@k+SimMix1,b,SAP2, RS@k2,c,SAP3, RS@k3},
        xtick={ContwM, Margin, RS@k0, RS@k+SimMix0, FastAP,BlackboxAP,  HORDE,ProxyNCAplus, ROADMAP, IntraBatch, CIA, SAP, SAP1, RS@k1, RS@k+SimMix1, SAP2, RS@k2,SAP3, RS@k3},
        xticklabels={Margin~\cite{wms+17}, Cont. w/M \cite{wzh+20},RS@k{\color{red}\textbf{$\dagger$}}, RS@k+SimMix{\color{red}\textbf{$\dagger$}}, FastAP~\cite{chx+19},Blackbox-AP~\cite{rmp+20}, HORDE~\cite{jph+19},ProxyNCA++~\cite{tdt20}, ROADMAP~\cite{ramzi2021robust}, IntraBatch~\cite{sel21}, CIA~\cite{kpm+23}, SAP~\cite{bxk+20}, SAP~\cite{bxk+20}{\color{red}\textbf{$\dagger$}}{\tiny +GeM+LN}, RS@k{\color{red}\textbf{$\dagger$}}, RS@k+SimMix{\color{red}\textbf{$\dagger$}}, SAP~\cite{bxk+20}{\color{red}\textbf{$\dagger$}}, RS@k{\color{red}\textbf{$\dagger$}}, SAP~\cite{bxk+20}{\color{red}\textbf{$\dagger$}}, RS@k{\color{red}\textbf{$\dagger$}} },
        title = {SOP},
        title style = {font=\tiny, yshift=-5pt},        
        x tick label style={rotate=45, align=right, text width=80pt, font=\tiny, xshift=-18pt, yshift=22pt},
        ylabel = {Recall@1},
        legend pos=north west,legend style={legend columns=-1, font=\footnotesize},
        ymin=60,ymax=100,
        bar width=0.5cm,                       
        ybar=-0.5cm,                           
        enlarge x limits={abs=0.9cm},
        nodes near coords,                     
        nodes near coords align={vertical},    
        every node near coord/.append style={font=\tiny,scale=1},        
        grid=none,
        ymajorgrids=true,xmajorticks=true,xminorticks=false,yminorgrids=true,
        legend image code/.code={\draw [#1] (0cm,-0.1cm) rectangle (0.3cm,0.35cm); },
]
        %
        \addplot[style={black,fill=Yellow,mark=none,postaction={pattern=grid, pattern color=black}}] coordinates {(Margin, 72.7)};
        \addplot[style={black,fill=RedOrange,mark=none,postaction={pattern=grid, pattern color=black}}] coordinates {(ContwM, 80.6)};                    
        \addplot[style={black,fill=LimeGreen,mark=none,postaction={pattern=grid, pattern color=black}}, nodes near coords=80.6] coordinates {(RS@k0,80.6)};
        \addplot[style={black,fill=Green,mark=none,postaction={pattern=grid, pattern color=black}}] coordinates {(RS@k+SimMix0,80.9)};
        \addplot[style={black,fill=Mulberry,mark=none,postaction={pattern=dots, pattern color=black}}] coordinates {(FastAP, 76.4)};            
        \addplot[style={black,fill=Salmon,mark=none,postaction={pattern=dots, pattern color=black}}] coordinates {(BlackboxAP, 78.6)};
        \addplot[style={black,fill=BrickRed,mark=none,postaction={pattern=dots, pattern color=black}}] coordinates {(HORDE, 80.1)};            
        \addplot[style={black,fill=GreenYellow,mark=none,postaction={pattern=dots, pattern color=black}}] coordinates {(ProxyNCAplus, 80.7)};        
        \addplot[style={black,fill=RubineRed,mark=none,postaction={pattern=dots, pattern color=black}}] coordinates {(ROADMAP,83.1)};
        \addplot[style={black,fill=BurntOrange,mark=none,postaction={pattern=dots, pattern color=black}}] coordinates {(IntraBatch, 81.4)};
        \addplot[style={black,fill=Red,mark=none,postaction={pattern=dots, pattern color=black}}] coordinates {(CIA, 82.3)};        
        \addplot[style={black,fill=Blue,mark=none,postaction={pattern=dots, pattern color=black}}] coordinates {(SAP,80.1)};
        \addplot[style={black,fill=SkyBlue,mark=none,postaction={pattern=dots, pattern color=black}}] coordinates {(SAP1,80.3)};
        \addplot[style={black,fill=LimeGreen,mark=none,postaction={pattern=dots, pattern color=black}}, nodes near coords=82.8] coordinates {(RS@k1,82.8)};
        \addplot[style={black,fill=Green,mark=none,postaction={pattern=dots, pattern color=black}}] coordinates {(RS@k+SimMix1,82.1)};        
        \addplot[style={black,fill=SkyBlue,mark=none,postaction={pattern=north east lines, pattern color=black}}, nodes near coords=83.7] coordinates {(SAP2,83.7)};
        \addplot[style={black,fill=LimeGreen,mark=none,postaction={pattern=north east lines, pattern color=black}}] coordinates {(RS@k2,85.1)};        
        \addplot[style={black,fill=SkyBlue,mark=none,postaction={pattern=crosshatch, pattern color=black}}, nodes near coords=86.6] coordinates {(SAP3,86.6)};
        \addplot[style={black,fill=LimeGreen,mark=none,postaction={pattern=crosshatch, pattern color=black}}] coordinates {(RS@k3,88.0)};        
    \end{axis}
\end{tikzpicture}
\\[-35pt]
\begin{tikzpicture}
    \begin{axis} [
        width  = \linewidth,
        height = 3.2cm,
        font=\footnotesize,
        symbolic x coords={ContwM, RS@k0, RS@k+SimMix0, a, FastAP,BlackboxAP, SAP, SAP1, RS@k1, RS@k+SimMix1,b,SAP2, RS@k2,c,SAP3, RS@k3},
        xtick={ContwM,RS@k0, RS@k+SimMix0, FastAP,BlackboxAP, SAP, SAP1, RS@k1, RS@k+SimMix1, SAP2, RS@k2,SAP3, RS@k3},
        xticklabels={Cont. w/M \cite{wzh+20}, RS@k{\color{red}\textbf{$\dagger$}}, RS@k+SimMix{\color{red}\textbf{$\dagger$}}, FastAP~\cite{chx+19},Blackbox-AP~\cite{rmp+20}, SAP~\cite{bxk+20}, SAP~\cite{bxk+20}{\color{red}\textbf{$\dagger$}}{\tiny +GeM+LN}, RS@k{\color{red}\textbf{$\dagger$}}, RS@k+SimMix{\color{red}\textbf{$\dagger$}}, SAP~\cite{bxk+20}{\color{red}\textbf{$\dagger$}}, RS@k{\color{red}\textbf{$\dagger$}}, SAP~\cite{bxk+20}{\color{red}\textbf{$\dagger$}}, RS@k{\color{red}\textbf{$\dagger$}} },
        title = {VehicleID - Large},
        title style = {font=\tiny, yshift=-5pt},        
        x tick label style={rotate=45, align=right, text width=80pt, font=\tiny, xshift=-18pt, yshift=22pt},
        ylabel = {Recall@1},
        legend pos=north west,legend style={legend columns=-1, font=\footnotesize},
        ymin=80,ymax=100,
        bar width=0.6cm,                       
        ybar=-0.6cm,                           
        enlarge x limits={abs=0.9cm},
        nodes near coords,                     
        nodes near coords align={vertical},    
        every node near coord/.append style={font=\tiny,scale=1},        
        grid=none,
        ymajorgrids=true,xmajorticks=true,xminorticks=false,yminorgrids=true,
        legend image code/.code={\draw [#1] (0cm,-0.1cm) rectangle (0.3cm,0.35cm); },
]
        %
        %
        \addplot[style={black,fill=RedOrange,mark=none,postaction={pattern=grid, pattern color=black}}] coordinates {(ContwM, 93.0)};
        \addplot[style={black,fill=LimeGreen,mark=none,postaction={pattern=grid, pattern color=black}}, nodes near coords=93.5] coordinates {(RS@k0,93.5)};
        \addplot[style={black,fill=Green,mark=none,postaction={pattern=grid, pattern color=black}}] coordinates {(RS@k+SimMix0,93.0)};
        \addplot[style={black,fill=Mulberry,mark=none,postaction={pattern=dots, pattern color=black}}] coordinates {(FastAP, 87.5)};            
        \addplot[style={black,fill=Salmon,mark=none,postaction={pattern=dots, pattern color=black}}] coordinates {(BlackboxAP, 93.0)};
        \addplot[style={black,fill=Blue,mark=none,postaction={pattern=dots, pattern color=black}}] coordinates {(SAP,91.9)};
        \addplot[style={black,fill=SkyBlue,mark=none,postaction={pattern=dots, pattern color=black}}] coordinates {(SAP1,91.0)};
        \addplot[style={black,fill=LimeGreen,mark=none,postaction={pattern=dots, pattern color=black}}, nodes near coords=93.8] coordinates {(RS@k1,93.8)};
        \addplot[style={black,fill=Green,mark=none,postaction={pattern=dots, pattern color=black}}] coordinates {(RS@k+SimMix1,93.3)};        
        \addplot[style={black,fill=SkyBlue,mark=none,postaction={pattern=north east lines, pattern color=black}}, nodes near coords=92.1] coordinates {(SAP2,92.1)};
        \addplot[style={black,fill=LimeGreen,mark=none,postaction={pattern=north east lines, pattern color=black}}] coordinates {(RS@k2,93.2)};        
        \addplot[style={black,fill=SkyBlue,mark=none,postaction={pattern=crosshatch, pattern color=black}}, nodes near coords=93.1] coordinates {(SAP3,93.1)};
        \addplot[style={black,fill=LimeGreen,mark=none,postaction={pattern=crosshatch, pattern color=black}}] coordinates {(RS@k3,94.7)};        
    \end{axis}
\end{tikzpicture}
\\[-35pt]
\begin{tikzpicture}
    \begin{axis} [
        width  = \linewidth,
        height = 3.2cm,
        font=\footnotesize,
        symbolic x coords={Margin,RS@k0, RS@k+SimMix0, a, HORDE,ProxyNCAplus, IntraBatch, CIA, SAP, SAP1, RS@k1, RS@k+SimMix1,b,SAP2, RS@k2,c,SAP3, RS@k3},
        xtick={Margin, RS@k0, RS@k+SimMix0, HORDE,ProxyNCAplus, IntraBatch, CIA, SAP, SAP1, RS@k1, RS@k+SimMix1, SAP2, RS@k2,SAP3, RS@k3},
        xticklabels={Margin~\cite{wms+17}, RS@k{\color{red}\textbf{$\dagger$}}, RS@k+SimMix{\color{red}\textbf{$\dagger$}}, HORDE~\cite{jph+19},ProxyNCA++~\cite{tdt20},  IntraBatch~\cite{sel21}, CIA~\cite{kpm+23}, SAP~\cite{bxk+20}, SAP~\cite{bxk+20}{\color{red}\textbf{$\dagger$}}{\tiny +GeM+LN}, RS@k{\color{red}\textbf{$\dagger$}}, RS@k+SimMix{\color{red}\textbf{$\dagger$}}, SAP~\cite{bxk+20}{\color{red}\textbf{$\dagger$}}, RS@k{\color{red}\textbf{$\dagger$}}, SAP~\cite{bxk+20}{\color{red}\textbf{$\dagger$}}, RS@k{\color{red}\textbf{$\dagger$}} },
        title = {Cars196},
        title style = {font=\tiny, yshift=-5pt},        
        x tick label style={rotate=45, align=right, text width=80pt, font=\tiny, xshift=-18pt, yshift=22pt},
        ylabel = {Recall@1},
        legend pos=north west,legend style={legend columns=-1, font=\footnotesize},
        ymin=70,ymax=100,
        bar width=0.6cm,                       
        ybar=-0.6cm,                           
        enlarge x limits={abs=0.9cm},
        nodes near coords,                     
        nodes near coords align={vertical},    
        every node near coord/.append style={font=\tiny,scale=1},        
        grid=none,
        ymajorgrids=true,xmajorticks=true,xminorticks=false,yminorgrids=true,
        legend image code/.code={\draw [#1] (0cm,-0.1cm) rectangle (0.3cm,0.35cm); },
]
        %
        \addplot[style={black,fill=Yellow,mark=none,postaction={pattern=grid, pattern color=black}}] coordinates {(Margin, 79.6)};
        \addplot[style={black,fill=LimeGreen,mark=none,postaction={pattern=grid, pattern color=black}}, nodes near coords=78.1] coordinates {(RS@k0,78.1)};
        \addplot[style={black,fill=Green,mark=none,postaction={pattern=grid, pattern color=black}}] coordinates {(RS@k+SimMix0,84.7)};
        \addplot[style={black,fill=BrickRed,mark=none,postaction={pattern=dots, pattern color=black}}] coordinates {(HORDE, 86.2)};            
        \addplot[style={black,fill=GreenYellow,mark=none,postaction={pattern=dots, pattern color=black}}] coordinates {(ProxyNCAplus, 86.5)};
        \addplot[style={black,fill=BurntOrange,mark=none,postaction={pattern=dots, pattern color=black}}] coordinates {(IntraBatch, 88.1)};
        \addplot[style={black,fill=Red,mark=none,postaction={pattern=dots, pattern color=black}}] coordinates {(CIA, 91.2)};                
        \addplot[style={black,fill=Blue,mark=none,postaction={pattern=dots, pattern color=black}}] coordinates {(SAP,76.1)};
        \addplot[style={black,fill=SkyBlue,mark=none,postaction={pattern=dots, pattern color=black}}] coordinates {(SAP1,78.2)};
        \addplot[style={black,fill=LimeGreen,mark=none,postaction={pattern=dots, pattern color=black}}, nodes near coords=80.7] coordinates {(RS@k1,80.7)};
        \addplot[style={black,fill=Green,mark=none,postaction={pattern=dots, pattern color=black}}] coordinates {(RS@k+SimMix1,88.2)};        
        \addplot[style={black,fill=SkyBlue,mark=none,postaction={pattern=north east lines, pattern color=black}}, nodes near coords=78.1] coordinates {(SAP2,78.1)};
        \addplot[style={black,fill=LimeGreen,mark=none,postaction={pattern=north east lines, pattern color=black}}] coordinates {(RS@k2,78.1)};        
        \addplot[style={black,fill=SkyBlue,mark=none,postaction={pattern=crosshatch, pattern color=black}}, nodes near coords=86.2] coordinates {(SAP3,86.2)};
        \addplot[style={black,fill=LimeGreen,mark=none,postaction={pattern=crosshatch, pattern color=black}}] coordinates {(RS@k3,89.5)};        
    \end{axis}
\end{tikzpicture}
\\[-20pt]
\end{tabular}

%% file: results_barplot_init.tex
\pgfplotsset{
    imgnet/.style={},
    dino/.style={postaction={pattern=north east lines, pattern color=black}},
    clipm/.style={postaction={pattern=grid, pattern color=black}},
    clipb/.style={postaction={pattern=crosshatch, pattern color=black}},
    diht/.style={postaction={pattern=dots, pattern color=black}},
    swag/.style={postaction={pattern=horizontal lines, pattern color=black}},
}
\begin{tikzpicture}
    \begin{axis} [
        font=\footnotesize,
        width  = 0.3\linewidth,
        height = 4.0cm,
        title = {iNaturalist - ViT-B/32},
        title style = {font=\tiny, yshift=-5pt},
        ylabel = {Recall@1},
        ylabel style={yshift=-5pt},
        symbolic x coords={init1,SAP1,RS@k1, a, init2,SAP2,RS@k2, b, init3,SAP3,RS@k3, c, init4,SAP4,RS@k4},
        xtick={SAP1,SAP2,SAP3,SAP4},
        xticklabels={IN-21k, CLIP-400m, CLIP-2B, DiHT},
        x tick label style={rotate=45, font=\tiny, yshift=5pt},
        legend pos=north west,legend style={legend columns=-1, font=\footnotesize},
        ymin=30,ymax=105,
        bar width=0.15cm,
        ybar=-.15cm,
        nodes near coords,     
        every node near coord/.append style={rotate=45,font=\tiny,scale=0.5, yshift=3pt, xshift=4pt}, 
        grid=none,
        ymajorgrids=true,xmajorticks=true,xminorticks=false,yminorgrids=true,
        legend image code/.code={\draw [#1] (0cm,-0.05cm) rectangle (0.1cm,0.15cm); },
]
        \addplot[style={mark=none,black,fill=Melon},imgnet] coordinates {(init1,46.9)};
        \addlegendentry{init}
        \addplot[style={mark=none,black,fill=SkyBlue},imgnet] coordinates {(SAP1,72.2)};
        \addlegendentry{SAP}
        \addplot[style={black,fill=LimeGreen,mark=none},imgnet] coordinates {(RS@k1,75.9)};        
        \addlegendentry{RS@k}
        \addplot[style={black,fill=Melon,mark=none},clipm] coordinates {(init2,53.9)};
        \addplot[style={black,fill=SkyBlue,mark=none},clipm] coordinates {(SAP2,62.6)};
        \addplot[style={black,fill=LimeGreen,mark=none},clipm] coordinates {(RS@k2,70.4)};        
        \addplot[style={black,fill=Melon,mark=none},clipb] coordinates {(init3,56.2)};
        \addplot[style={black,fill=SkyBlue,mark=none},clipb] coordinates {(SAP3,62.9)};
        \addplot[style={black,fill=LimeGreen,mark=none},clipb] coordinates {(RS@k3,71.2)};        
        \addplot[style={black,fill=Melon,mark=none},diht] coordinates {(init4,58.5)};
        \addplot[style={black,fill=SkyBlue,mark=none},diht] coordinates {(SAP4,66.8)};
        \addplot[style={black,fill=LimeGreen,mark=none},diht] coordinates {(RS@k4,73.3)};        
    \end{axis}
\end{tikzpicture}
\hspace{-10pt}
\begin{tikzpicture}
    \begin{axis} [
        font=\footnotesize,
        width  = 0.37\linewidth,
        height = 4.0cm,
        title = {iNaturalist - ViT-B/16},
        title style = {font=\tiny, yshift=-5pt},
        yticklabels=none,        
        symbolic x coords={init1,SAP1,RS@k1, a, init2,SAP2,RS@k2, b, init3,SAP3,RS@k3, c, init4,SAP4,RS@k4, d, init5,SAP5,RS@k5},
        xtick={SAP1, SAP2, SAP3, SAP4, SAP5},
        xticklabels={IN-21k, CLIP-400m, CLIP-2B, DiHT, SWAG},
        x tick label style={rotate=45, font=\tiny, yshift=5pt},
        ymin=30,ymax=105,
        bar width=0.15cm,
        ybar=-.15cm,
        nodes near coords,     
        every node near coord/.append style={rotate=45,font=\tiny,scale=0.5, yshift=3pt, xshift=4pt}, 
        grid=none,
        ymajorgrids=true,xmajorticks=true,xminorticks=false,yminorgrids=true,
]
        \addplot[style={black,fill=Melon,mark=none},imgnet] coordinates {(init1,59.3)};
        \addplot[style={black,fill=SkyBlue,mark=none},imgnet] coordinates {(SAP1,79.1)};
        \addplot[style={black,fill=LimeGreen,mark=none},imgnet] coordinates {(RS@k1,83.9)};        
        \addplot[style={black,fill=Melon,mark=none},clipm] coordinates {(init2,62.9)};
        \addplot[style={black,fill=SkyBlue,mark=none},clipm] coordinates {(SAP2,73.7)};
        \addplot[style={black,fill=LimeGreen,mark=none},clipm] coordinates {(RS@k2,80.1)};        
        \addplot[style={black,fill=Melon,mark=none},clipb] coordinates {(init3,64.4)};
        \addplot[style={black,fill=SkyBlue,mark=none},clipb] coordinates {(SAP3,74.2)};
        \addplot[style={black,fill=LimeGreen,mark=none},clipb] coordinates {(RS@k3,80.3)};        
        \addplot[style={black,fill=Melon,mark=none},diht] coordinates {(init4,67.2)};
        \addplot[style={black,fill=SkyBlue,mark=none},diht] coordinates {(SAP4,76.8)};
        \addplot[style={black,fill=LimeGreen,mark=none},diht] coordinates {(RS@k4,82.4)};        
        \addplot[style={black,fill=Melon,mark=none},swag] coordinates {(init5,73.1)};
        \addplot[style={black,fill=SkyBlue,mark=none},swag] coordinates {(SAP5,83.5)};
        \addplot[style={black,fill=LimeGreen,mark=none},swag] coordinates {(RS@k5,84.9)};        
    \end{axis}
\end{tikzpicture}
\hspace{-10pt}
\begin{tikzpicture}
    \begin{axis} [
        font=\footnotesize,
        width  = 0.23\linewidth,
        height = 4.0cm,
        title = {iNaturalist - ViT-L/16},
        title style = {font=\tiny, yshift=-5pt},
        yticklabels=none,
        symbolic x coords={init1,SAP1,RS@k1, a, init2,SAP2,RS@k2},
        xtick={SAP1, SAP2},
        xticklabels={\hspace{9.5pt}IN-21k, SWAG},
        x tick label style={rotate=45, font=\tiny, yshift=5pt},
        ymin=30,ymax=105,
        bar width=0.15cm,
        ybar=-.15cm,
        nodes near coords,
        every node near coord/.append style={rotate=45,font=\tiny,scale=0.5, yshift=3pt, xshift=4pt},
        grid=none,
        ymajorgrids=true,xmajorticks=true,xminorticks=false,yminorgrids=true,
]
        \addplot[style={black,fill=Melon,mark=none},imgnet] coordinates {(init1,69.3)};
        \addplot[style={black,fill=SkyBlue,mark=none},imgnet] coordinates {(SAP1,82.6)};
        \addplot[style={black,fill=LimeGreen,mark=none},imgnet] coordinates {(RS@k1,84.9)};        
        \addplot[style={black,fill=Melon,mark=none},swag] coordinates {(init2,73.1)};
        \addplot[style={black,fill=SkyBlue,mark=none},swag] coordinates {(SAP2,83.5)};
        \addplot[style={black,fill=LimeGreen,mark=none},swag] coordinates {(RS@k2,84.9)};        
    \end{axis}
\end{tikzpicture}
\hspace{-10pt}
\begin{tikzpicture}
    \begin{axis} [
        font=\footnotesize,
        width  = 0.32\linewidth,
        height = 4.0cm,
        title = {iNaturalist - ViT-L/14},
        title style = {font=\tiny, yshift=-5pt},
        yticklabels=none,        
        symbolic x coords={init1,SAP1,RS@k1, a, init2,SAP2,RS@k2, b, init3,SAP3,RS@k3, c, init4,SAP4,RS@k4},
        xtick={SAP1, SAP2, SAP3, SAP4},
        xticklabels={CLIP-400m, CLIP-2B, DiHT, DINOv2},
        x tick label style={rotate=45, font=\tiny, yshift=5pt},
        ymin=30,ymax=105,
        bar width=0.15cm,
        ybar=-.15cm,
        nodes near coords,     
        every node near coord/.append style={rotate=45,font=\tiny,scale=0.5, yshift=3pt, xshift=4pt}, 
        grid=none,
        ymajorgrids=true,xmajorticks=true,xminorticks=false,yminorgrids=true,
]
        \addplot[style={black,fill=Melon,mark=none},clipm] coordinates {(init1,70.1)};
        \addplot[style={black,fill=SkyBlue,mark=none},clipm] coordinates {(SAP1,77.8)};
        \addplot[style={black,fill=LimeGreen,mark=none},clipm] coordinates {(RS@k1,82.6)};        
        \addplot[style={black,fill=Melon,mark=none},clipb] coordinates {(init2,66.0)};
        \addplot[style={black,fill=SkyBlue,mark=none},clipb] coordinates {(SAP2,77.9)};
        \addplot[style={black,fill=LimeGreen,mark=none},clipb] coordinates {(RS@k2,82.8)};        
        \addplot[style={black,fill=Melon,mark=none},diht] coordinates {(init3,74.9)};
        \addplot[style={black,fill=SkyBlue,mark=none},diht] coordinates {(SAP3,81.3)};
        \addplot[style={black,fill=LimeGreen,mark=none},diht] coordinates {(RS@k3,85.2)};        
        \addplot[style={black,fill=Melon,mark=none},dino] coordinates {(init4,82.2)};
        \addplot[style={black,fill=SkyBlue,mark=none},dino] coordinates {(SAP4,88.5)};
        \addplot[style={black,fill=LimeGreen,mark=none},dino] coordinates {(RS@k4,90.0)};        
    \end{axis}
\end{tikzpicture}
\begin{tikzpicture}
    \begin{axis} [
        font=\footnotesize,
        width  = 0.3\linewidth,
        height = 4.0cm,
        title = {SOP - ViT-B/32},
        title style = {font=\tiny, yshift=-5pt},
        ylabel = {Recall@1},
        ylabel style={yshift=-5pt},
        symbolic x coords={init1,SAP1,RS@k1, a, init2,SAP2,RS@k2, b, init3,SAP3,RS@k3, c, init4,SAP4,RS@k4},
        xtick={SAP1,SAP2,SAP3,SAP4},
        xticklabels={IN-21k, CLIP-400m, CLIP-2B, DiHT},
        x tick label style={rotate=45, font=\tiny, yshift=5pt},
        ymin=30,ymax=105,
        bar width=0.15cm,
        ybar=-.15cm,
        nodes near coords,     
        every node near coord/.append style={rotate=45,font=\tiny,scale=0.5, yshift=3pt, xshift=4pt}, 
        grid=none,
        ymajorgrids=true,xmajorticks=true,xminorticks=false,yminorgrids=true,
]
        \addplot[style={mark=none,black,fill=Melon},imgnet] coordinates {(init1,54.7)};
        \addplot[style={mark=none,black,fill=SkyBlue},imgnet] coordinates {(SAP1,83.7)};
        \addplot[style={black,fill=LimeGreen,mark=none},imgnet] coordinates {(RS@k1,85.1)};        
        %
        \addplot[style={black,fill=Melon,mark=none},clipm] coordinates {(init2,62.7)};
        \addplot[style={black,fill=SkyBlue,mark=none},clipm] coordinates {(SAP2,80.6)};
        \addplot[style={black,fill=LimeGreen,mark=none},clipm] coordinates {(RS@k2,84.3)};        
        \addplot[style={black,fill=Melon,mark=none},clipb] coordinates {(init3,64.1)};
        \addplot[style={black,fill=SkyBlue,mark=none},clipb] coordinates {(SAP3,82.3)};
        \addplot[style={black,fill=LimeGreen,mark=none},clipb] coordinates {(RS@k3,86.4)};        
        \addplot[style={black,fill=Melon,mark=none},diht] coordinates {(init4,62.2)};
        \addplot[style={black,fill=SkyBlue,mark=none},diht] coordinates {(SAP4,83.3)};
        \addplot[style={black,fill=LimeGreen,mark=none},diht] coordinates {(RS@k4,85.9)};        
    \end{axis}
\end{tikzpicture}
\hspace{-10pt}
\begin{tikzpicture}
    \begin{axis} [
        font=\footnotesize,
        width  = 0.37\linewidth,
        height = 4.0cm,
        title = {SOP - ViT-B/16},
        title style = {font=\tiny, yshift=-5pt},
        yticklabels=none,        
        symbolic x coords={init1,SAP1,RS@k1, a, init2,SAP2,RS@k2, b, init3,SAP3,RS@k3, c, init4,SAP4,RS@k4, d, init5,SAP5,RS@k5},
        xtick={SAP1, SAP2, SAP3, SAP4, SAP5},
        xticklabels={IN-21k, CLIP-400m, CLIP-2B, DiHT, SWAG},
        x tick label style={rotate=45, font=\tiny, yshift=5pt},
        ymin=30,ymax=105,
        bar width=0.15cm,
        ybar=-.15cm,
        nodes near coords,     
        every node near coord/.append style={rotate=45,font=\tiny,scale=0.5, yshift=3pt, xshift=4pt}, 
        grid=none,
        ymajorgrids=true,xmajorticks=true,xminorticks=false,yminorgrids=true,
]
        \addplot[style={black,fill=Melon,mark=none},imgnet] coordinates {(init1,55.8)};
        \addplot[style={black,fill=SkyBlue,mark=none},imgnet] coordinates {(SAP1,86.6)};
        \addplot[style={black,fill=LimeGreen,mark=none},imgnet] coordinates {(RS@k1,88.0)};        
        \addplot[style={black,fill=Melon,mark=none},clipm] coordinates {(init2,65.7)};
        \addplot[style={black,fill=SkyBlue,mark=none},clipm] coordinates {(SAP2,85.0)};
        \addplot[style={black,fill=LimeGreen,mark=none},clipm] coordinates {(RS@k2,87.7)};        
        \addplot[style={black,fill=Melon,mark=none},clipb] coordinates {(init3,67.0)};
        \addplot[style={black,fill=SkyBlue,mark=none},clipb] coordinates {(SAP3,86.6)};
        \addplot[style={black,fill=LimeGreen,mark=none},clipb] coordinates {(RS@k3,89.0)};        
        \addplot[style={black,fill=Melon,mark=none},diht] coordinates {(init4,65.6)};
        \addplot[style={black,fill=SkyBlue,mark=none},diht] coordinates {(SAP4,87.0)};
        \addplot[style={black,fill=LimeGreen,mark=none},diht] coordinates {(RS@k4,88.9)};        
        \addplot[style={black,fill=Melon,mark=none},swag] coordinates {(init5,59.3)};
        \addplot[style={black,fill=SkyBlue,mark=none},swag] coordinates {(SAP5,85.8)};
        \addplot[style={black,fill=LimeGreen,mark=none},swag] coordinates {(RS@k5,89.2)};        
    \end{axis}
\end{tikzpicture}
\hspace{-10pt}
\begin{tikzpicture}
    \begin{axis} [
        font=\footnotesize,
        width  = 0.23\linewidth,
        height = 4.0cm,
        title = {SOP - ViT-L/16},
        title style = {font=\tiny, yshift=-5pt},
        yticklabels=none,
        symbolic x coords={init1,SAP1,RS@k1, a, init2,SAP2,RS@k2},
        xtick={SAP1, SAP2},
        xticklabels={\hspace{9.5pt}IN-21k, SWAG},
        x tick label style={rotate=45, font=\tiny, yshift=5pt},
        ymin=30,ymax=105,
        bar width=0.15cm,
        ybar=-.15cm,
        nodes near coords,
        every node near coord/.append style={rotate=45,font=\tiny,scale=0.5, yshift=3pt, xshift=4pt},
        grid=none,
        ymajorgrids=true,xmajorticks=true,xminorticks=false,yminorgrids=true,
]
        \addplot[style={black,fill=Melon,mark=none},imgnet] coordinates {(init1,64.9)};
        \addplot[style={black,fill=SkyBlue,mark=none},imgnet] coordinates {(SAP1,86.6)};
        \addplot[style={black,fill=LimeGreen,mark=none},imgnet] coordinates {(RS@k1,88.6)};        
        \addplot[style={black,fill=Melon,mark=none},swag] coordinates {(init2,59.3)};
        \addplot[style={black,fill=SkyBlue,mark=none},swag] coordinates {(SAP2,85.8)};
        \addplot[style={black,fill=LimeGreen,mark=none},swag] coordinates {(RS@k2,89.2)};        
    \end{axis}
\end{tikzpicture}
\hspace{-10pt}
\begin{tikzpicture}
    \begin{axis} [
        font=\footnotesize,
        width  = 0.32\linewidth,
        height = 4.0cm,
        title = {SOP - ViT-L/14},
        title style = {font=\tiny, yshift=-5pt},
        yticklabels=none,        
        symbolic x coords={init1,SAP1,RS@k1, a, init2,SAP2,RS@k2, b, init3,SAP3,RS@k3, c, init4,SAP4,RS@k4},
        xtick={SAP1, SAP2, SAP3, SAP4},
        xticklabels={CLIP-400m, CLIP-2B, DiHT, DINOv2},
        x tick label style={rotate=45, font=\tiny, yshift=5pt},
        ymin=30,ymax=105,
        bar width=0.15cm,
        ybar=-.15cm,
        nodes near coords,     
        every node near coord/.append style={rotate=45,font=\tiny,scale=0.5, yshift=3pt, xshift=4pt}, 
        grid=none,
        ymajorgrids=true,xmajorticks=true,xminorticks=false,yminorgrids=true,
]
        \addplot[style={black,fill=Melon,mark=none},clipm] coordinates {(init1,67.2)};
        \addplot[style={black,fill=SkyBlue,mark=none},clipm] coordinates {(SAP1,86.2)};
        \addplot[style={black,fill=LimeGreen,mark=none},clipm] coordinates {(RS@k1,89.3)};        
        \addplot[style={black,fill=Melon,mark=none},clipb] coordinates {(init2,65.9)};
        \addplot[style={black,fill=SkyBlue,mark=none},clipb] coordinates {(SAP2,87.5)};
        \addplot[style={black,fill=LimeGreen,mark=none},clipb] coordinates {(RS@k2,90.6)};        
        \addplot[style={black,fill=Melon,mark=none},diht] coordinates {(init3,68.3)};
        \addplot[style={black,fill=SkyBlue,mark=none},diht] coordinates {(SAP3,87.4)};
        \addplot[style={black,fill=LimeGreen,mark=none},diht] coordinates {(RS@k3,90.8)};        
        \addplot[style={black,fill=Melon,mark=none},dino] coordinates {(init4,55.6)};
        \addplot[style={black,fill=SkyBlue,mark=none},dino] coordinates {(SAP4,87.5)    };
        \addplot[style={black,fill=LimeGreen,mark=none},dino] coordinates {(RS@k4,90.5)};        
    \end{axis}
\end{tikzpicture}
\begin{tikzpicture}
    \begin{axis} [
        font=\footnotesize,
        width  = 0.3\linewidth,
        height = 4.0cm,
        title = {Cars196 - ViT-B/32},
        title style = {font=\tiny, yshift=-5pt},
        ylabel = {Recall@1},
        ylabel style={yshift=-5pt},
        symbolic x coords={init1,SAP1,RS@k1, a, init2,SAP2,RS@k2, b, init3,SAP3,RS@k3, c, init4,SAP4,RS@k4},
        xtick={SAP1,SAP2,SAP3,SAP4},
        xticklabels={IN-21k, CLIP-400m, CLIP-2B, DiHT},
        x tick label style={rotate=45, font=\tiny, yshift=5pt},
        ymin=30,ymax=105,
        bar width=0.15cm,
        ybar=-.15cm,
        nodes near coords,     
        every node near coord/.append style={rotate=45,font=\tiny,scale=0.5, yshift=3pt, xshift=4pt}, 
        grid=none,
        ymajorgrids=true,xmajorticks=true,xminorticks=false,yminorgrids=true,
]
        \addplot[style={mark=none,black,fill=Melon},imgnet] coordinates {(init1,33.2)};
        \addplot[style={mark=none,black,fill=SkyBlue},imgnet] coordinates {(SAP1,78.1)};
        \addplot[style={black,fill=LimeGreen,mark=none},imgnet] coordinates {(RS@k1,78.1)};        
        %
        \addplot[style={black,fill=Melon,mark=none},clipm] coordinates {(init2,87.0)};
        \addplot[style={black,fill=SkyBlue,mark=none},clipm] coordinates {(SAP2,92.6)};
        \addplot[style={black,fill=LimeGreen,mark=none},clipm] coordinates {(RS@k2,91.8)};        
        \addplot[style={black,fill=Melon,mark=none},clipb] coordinates {(init3,90.1)};
        \addplot[style={black,fill=SkyBlue,mark=none},clipb] coordinates {(SAP3,94.4)};
        \addplot[style={black,fill=LimeGreen,mark=none},clipb] coordinates {(RS@k3,94.2)};        
        \addplot[style={black,fill=Melon,mark=none},diht] coordinates {(init4,87.7)};
        \addplot[style={black,fill=SkyBlue,mark=none},diht] coordinates {(SAP4,93.6)};
        \addplot[style={black,fill=LimeGreen,mark=none},diht] coordinates {(RS@k4,92.8)};        
    \end{axis}
\end{tikzpicture}
\hspace{-10pt}
\begin{tikzpicture}
    \begin{axis} [
        font=\footnotesize,
        width  = 0.37\linewidth,
        height = 4.0cm,
        title = {Cars196 - ViT-B/16},
        title style = {font=\tiny, yshift=-5pt},
        yticklabels=none,        
        symbolic x coords={init1,SAP1,RS@k1, a, init2,SAP2,RS@k2, b, init3,SAP3,RS@k3, c, init4,SAP4,RS@k4, d, init5,SAP5,RS@k5},
        xtick={SAP1, SAP2, SAP3, SAP4, SAP5},
        xticklabels={IN-21k, CLIP-400m, CLIP-2B, DiHT, SWAG},
        x tick label style={rotate=45, font=\tiny, yshift=5pt},
        ymin=30,ymax=105,
        bar width=0.15cm,
        ybar=-.15cm,
        nodes near coords,     
        every node near coord/.append style={rotate=45,font=\tiny,scale=0.5, yshift=3pt, xshift=4pt}, 
        grid=none,
        ymajorgrids=true,xmajorticks=true,xminorticks=false,yminorgrids=true,
]
        \addplot[style={black,fill=Melon,mark=none},imgnet] coordinates {(init1,35.6)};
        \addplot[style={black,fill=SkyBlue,mark=none},imgnet] coordinates {(SAP1,86.2)};
        \addplot[style={black,fill=LimeGreen,mark=none},imgnet] coordinates {(RS@k1,89.5)};        
        \addplot[style={black,fill=Melon,mark=none},clipm] coordinates {(init2,91.4)};
        \addplot[style={black,fill=SkyBlue,mark=none},clipm] coordinates {(SAP2,95.2)};
        \addplot[style={black,fill=LimeGreen,mark=none},clipm] coordinates {(RS@k2,95.5)};        
        \addplot[style={black,fill=Melon,mark=none},clipb] coordinates {(init3,92.7)};
        \addplot[style={black,fill=SkyBlue,mark=none},clipb] coordinates {(SAP3,96.5)};
        \addplot[style={black,fill=LimeGreen,mark=none},clipb] coordinates {(RS@k3,96.4)};        
        \addplot[style={black,fill=Melon,mark=none},diht] coordinates {(init4,91.1)};
        \addplot[style={black,fill=SkyBlue,mark=none},diht] coordinates {(SAP4,96.1)};
        \addplot[style={black,fill=LimeGreen,mark=none},diht] coordinates {(RS@k4,96.2)};        
        \addplot[style={black,fill=Melon,mark=none},swag] coordinates {(init5,76.5)};
        \addplot[style={black,fill=SkyBlue,mark=none},swag] coordinates {(SAP5,92.8)};
        \addplot[style={black,fill=LimeGreen,mark=none},swag] coordinates {(RS@k5,92.5)};        
    \end{axis}
\end{tikzpicture}
\hspace{-10pt}
\begin{tikzpicture}
    \begin{axis} [
        font=\footnotesize,
        width  = 0.23\linewidth,
        height = 4.0cm,
        title = {Cars196 - ViT-L/16},
        title style = {font=\tiny, yshift=-5pt},
        yticklabels=none,
        symbolic x coords={init1,SAP1,RS@k1, a, init2,SAP2,RS@k2},
        xtick={SAP1, SAP2},
        xticklabels={\hspace{9.5pt}IN-21k, SWAG},
        x tick label style={rotate=45, font=\tiny, yshift=5pt},
        ymin=30,ymax=105,
        bar width=0.15cm,
        ybar=-.15cm,
        nodes near coords,
        every node near coord/.append style={rotate=45,font=\tiny,scale=0.5, yshift=3pt, xshift=4pt},
        grid=none,
        ymajorgrids=true,xmajorticks=true,xminorticks=false,yminorgrids=true,
]
        \addplot[style={black,fill=Melon,mark=none},imgnet] coordinates {(init1,50.1)};
        \addplot[style={black,fill=SkyBlue,mark=none},imgnet] coordinates {(SAP1,91.3)};
        \addplot[style={black,fill=LimeGreen,mark=none},imgnet] coordinates {(RS@k1,90.4)};        
        \addplot[style={black,fill=Melon,mark=none},swag] coordinates {(init2,81.1)};
        \addplot[style={black,fill=SkyBlue,mark=none},swag] coordinates {(SAP2,95.0)};
        \addplot[style={black,fill=LimeGreen,mark=none},swag] coordinates {(RS@k2,94.8)};        
    \end{axis}
\end{tikzpicture}
\hspace{-10pt}
\begin{tikzpicture}
    \begin{axis} [
        font=\footnotesize,
        width  = 0.32\linewidth,
        height = 4.0cm,
        title = {Cars196 - ViT-L/14},
        title style = {font=\tiny, yshift=-5pt},
        yticklabels=none,        
        symbolic x coords={init1,SAP1,RS@k1, a, init2,SAP2,RS@k2, b, init3,SAP3,RS@k3, c, init4,SAP4,RS@k4},
        xtick={SAP1, SAP2, SAP3, SAP4},
        xticklabels={CLIP-400m, CLIP-2B, DiHT, DINOv2},
        x tick label style={rotate=45, font=\tiny, yshift=5pt},
        ymin=30,ymax=105,
        bar width=0.15cm,
        ybar=-.15cm,
        nodes near coords,     
        every node near coord/.append style={rotate=45,font=\tiny,scale=0.5, yshift=3pt, xshift=4pt}, 
        grid=none,
        ymajorgrids=true,xmajorticks=true,xminorticks=false,yminorgrids=true,
]
        \addplot[style={black,fill=Melon,mark=none},clipm] coordinates {(init1,93.4)};
        \addplot[style={black,fill=SkyBlue,mark=none},clipm] coordinates {(SAP1,96.6)};
        \addplot[style={black,fill=LimeGreen,mark=none},clipm] coordinates {(RS@k1,96.7)};        
        \addplot[style={black,fill=Melon,mark=none},clipb] coordinates {(init2,93.8)};
        \addplot[style={black,fill=SkyBlue,mark=none},clipb] coordinates {(SAP2,96.9)};
        \addplot[style={black,fill=LimeGreen,mark=none},clipb] coordinates {(RS@k2,97.0)};        
        \addplot[style={black,fill=Melon,mark=none},diht] coordinates {(init3,93.9)};
        \addplot[style={black,fill=SkyBlue,mark=none},diht] coordinates {(SAP3,97.1)};
        \addplot[style={black,fill=LimeGreen,mark=none},diht] coordinates {(RS@k3,97.2)};        
        \addplot[style={black,fill=Melon,mark=none},dino] coordinates {(init4,82.9)};
        \addplot[style={black,fill=SkyBlue,mark=none},dino] coordinates {(SAP4,96.3)};
        \addplot[style={black,fill=LimeGreen,mark=none},dino] coordinates {(RS@k4,96.8)};        
    \end{axis}
\end{tikzpicture}

%% file: tab_rop.tex
%
\setlength\extrarowheight{0pt}
\newcommand{\xdagger}{^{\dagger}}
\def\arraystretch{1.0}
\small
\begin{tabular}	{l@{\msp}r@{\msp}r@{\msp}r@{\msp}c@{\msp}c@{\msp}c@{\msp}c@{\msp}c@{\msp}c@{\msp}c@{\msp}c@{\msp}c@{\msp}c}
\hline
\multirow{2}{*}{Arch.}        &    \multirow{2}{*}{Loss} & \multirow{2}{*}{Train-set} & \multirow{2}{*}{} & \multicolumn{2}{c}{Mean}  &   \multicolumn{2}{c}{\ro} & \multicolumn{2}{c}{\ro\hspace{-3pt}+\rdis} & \multicolumn{2}{c}{\rpa} & \multicolumn{2}{c}{\rp\hspace{-3pt}+\rdis}     \\ \cline{5-14}
															 & &&    												     & all & \rdis &   med  &   hard         &      med  &   hard          &    med  &   hard      &           med  &   hard         \\ \hline \hline
GeM$\ast$                                         & AP \cite{hls18} & Landmarks-clean~\cite{bsc+14}\cite{gar+17} & ~\cite{rar+19}/~\cite{tjc20}    & 49.7 & 36.7 & 67.1 & 42.3 & 47.8 & 22.5 & 80.3 & 60.9 & 51.9 & 24.6 \\             
GeM$\ast$                                         & AP \cite{hls18} & GLDv1~\cite{nas+17} & ~\cite{rar+19}/github    & - & -  & 66.3 & 42.5 & - & -  & 80.2 & 60.8  & -  & - \\             
GeM                & SAP \cite{bxk+20} & GLDv1~\cite{nas+17} & ~\cite{bxk+20}    & 52.7 & 40.6 & 67.9 & 46.3 &  49.5 & 25.8 & 81.7 & 63.3 & 57.4 & 29.8 \\     
GeM                & RS@k & GLDv1~\cite{nas+17} & ours    & 53.1 & 41.0 & 68.3 & 46.1 & 50.1 & 25.8 & 82.1 & 63.9 & 57.9 & 30.2 \\            
GeM+SiMix                & RS@k & GLDv1~\cite{nas+17} & ours    & 53.1 & 41.8 & 68.4 & 45.3 & 51.0 & 26.4 & 81.2 & 62.4 & 58.7 & 31.1\\            
 \hline
\end{tabular}

%% file: fig_temp.tex
\input{data_ab_temp}
\begin{tikzpicture}
\begin{axis}[%
	width=0.8\linewidth,
	height=0.6\linewidth,
	xlabel={\small $\tau_{1}$},
	ylabel={\small r@1},
	title={varying temperature},
	legend cell align={left},
	legend pos=south east,
    legend style={cells={anchor=east}, font =\tiny, fill opacity=0.7, row sep=-2.5pt},
   	xtick={0.1,0.5,1.0,2.0,5.0},
   	xticklabels={0.1,0.5,1.0,2.0,5.0},
   	xmode=log,
   	ymin=70,
    grid=both,
]
	\addplot[color=magenta,     solid, mark=*,  mark size=1.5, line width=1.0] table[x=temp, y expr={\thisrow{cars_r1}}] 
	\yfccLambda;
	\addlegendentry{RS@k};
	\addplot[color=magenta,     dashed, mark=*,  mark size=1.5, line width=1.0] table[x=temp, y expr={\thisrow{cars_simix_r1}}] \yfccLambda;
	\addlegendentry{RS@k+SiMix};
\end{axis}
\end{tikzpicture}

%% file: fig_bs.tex
\input{data_ab_bs}
\begin{tikzpicture}
\begin{axis}[%
	width=0.8\linewidth,
	height=0.6\linewidth,
	xlabel={\small batch size},
	ylabel={\small r@1},
	title={varying batch size},
	legend cell align={left},
	legend pos=south east,
    legend style={cells={anchor=east}, font =\tiny, fill opacity=0.7, row sep=-2.5pt},
   	xtick={16,32,64,128,256,512,1024,2048,4096},
   	xmode=log,
   	log basis x={2},
    grid=both,
]
	\addplot[color=blue,     solid, mark=*,  mark size=1.5, line width=1.0] table[x=bs_cars, y expr={\thisrow{cars_r1_sap}}]
	\yfccLambda;
	\addlegendentry{SAP};
	\addplot[color=magenta,     solid, mark=*,  mark size=1.5, line width=1.0] table[x=bs_cars, y expr={\thisrow{cars_r1}}] 
	\yfccLambda;
	\addlegendentry{RS@k};
	\addplot[color=magenta,     dashed, mark=*,  mark size=1.5, line width=1.0] table[x=bs_cars, y expr={\thisrow{cars_simix_r1}}]
	\yfccLambda;
	\addlegendentry{RS@k+SiMix};
\end{axis}
\end{tikzpicture}

%% file: data_ab_bs.tex
\pgfplotstableread{
 		bs_sop	sop_r1     bs_cars     cars_r1      cars_simix_r1   cars_r1_sap
 		16		72.9	    16          74.60       86.27           68.9
 		32		77.3	    32          77.04       87.46           71.91
 		64		79.0	    64          76.67       87.31           75.64
 		128		79.7	    128         78.92       87.43           77.92
 		256		80.5        256         79.14       87.06           78.48
 		512     81.2        392         80.72       88.17           79.50
 		1024    82.0        nan         nan         nan             nan
 		2048    82.5        nan         nan         nan             nan
 		4096    82.8        nan         nan         nan             nan
 		8192    82.4        nan         nan         nan             nan
 	}{\yfccLambda}

%% file: tab_cars196_ab.tex
\setlength\extrarowheight{0pt}
   \begin{tabular}{l@{\msp}c@{\msp}c@{\msp}c@{\msp}c@{\msp}c@{\msp}c}
    \hline
    \multicolumn{1}{l}{Method} &  r@1   & r@2 &  r@4  & r@8 & r@16 & Avg\\
    	\hline\hline
    	
        RS@$\{1\}$ &
        $81.1$ &
        $87.7$ &
        $92.0$ &
        $95.0$ &
        $96.9$ &
        $90.5$
        \\
        
        RS@$\{1,2\}$ &
        $80.2$ &
        $87.2$ &
        $91.9$ &
        $95.0$ &
        $97.2$ &
        $90.3$
        \\
	
	    RS@$\{1,2,4\}$ &
        $79.6$ &
        $86.5$ &
        $91.2$ &
        $94.5$ &
        $96.8$ &
        $89.7$
        \\
        
        RS@$\{1,2,4,8\}$ &
        $79.3$ &
        $86.3$ &
        $91.0$ &
        $94.5$ &
        $96.9$ &
        $89.6$
        \\
        
        RS@$\{1,2,4,8,16\}$ &
        $80.8$ &
        $87.6$ &
        $92.2$ &
        $95.0$ &
        $97.1$ &
        $90.5$
        \\
        
        RS@$\{2,4,8,16\}$ &
        $80.3$ &
        $87.5$ &
        $92.3$ &
        $95.4$ &
        $97.5$ &
        $90.6$
        \\
        
        RS@$\{4,8,16\}$ &
        $79.6$ &
        $87.1$ &
        $91.7$ &
        $95.0$ &
        $97.3$ &
        $90.1$
        \\
        RS@$\{8,16\}$ &
        $79.6$ &
        $87.1$ &
        $91.7$ &
        $95.0$ &
        $97.3$ &
        $90.1$
        \\
        RS@$\{16\}$ &
        $75.8$ & 
        $83.9$ &
        $89.8$ &
        $93.6$ &
        $96.4$ &
        $87.9$
        \\
	\hline
    \end{tabular}

%% file: tab_metriclearning.tex
    \begin{tabular}{@{\zsp}l@{\xssp}l@{\xssp}@{\ssp}cccc@{\ssp}cccc@{\ssp}cc@{\ssp}cc@{\ssp}cc@{\ssp}cccc}
    \hline
    \multirow{2}[1]{*}{Method}& \multirow{2}[1]{*}{Arch.$^\text{dim}$} & \multicolumn{4}{c}{iNaturalist \cite{vms+18}} & \multicolumn{4}{c}{SOP \cite{ohb16}} & \multicolumn{6}{c}{VehicleID \cite{ltw+16}} & \multicolumn{4}{c}{Cars196 \cite{ksd+13}}\\
    \cline{3-20}
    & & & & & & & & & & \multicolumn{2}{c}{Small} & \multicolumn{2}{c}{Medium} & \multicolumn{2}{c}{Large} & & & & \\
    & & 1 & 4 & 16 & 32 & $10^0$ & $10^1$ & $10^2$ & $10^3$ & 1 & 5 & 1 & 5 & 1 & 5 & 1 & 2 & 4 & 8\\
    \hline \hline
    
	ProxyNCA \cite{mtl+17} & {\scriptsize$I_{1}^{128}$} &
	$\underline{61.6}$ & 
	$\underline{77.4}$ & 
	$\underline{87.0}$ & 
	$90.6$ &

	$73.7$ & 
	- & 
	- & 
	- &
	
	- &
	- &
	- &
	- &
	- &
	- &
	
	$73.2$ &
	$82.4$ &
	$86.4$ &
	$88.7$
	\\
	
	Margin \cite{wms+17} & {\scriptsize$R_{50}^{128}$} &
	$58.1$ & 
	$75.5$ & 
	$86.8$ & 
	$\underline{90.7}$ &

	$72.7$ &
	$86.2$ & 
	$93.8$ & 
	$98.0$ &
	
	- &
	- &
	- &
	- &
	- &
	- &
	
	$\underline{79.6}$ &
	$\underline{86.5}$ &
	$\underline{91.9}$ &
	$\underline{95.1}$
	\\
	
	Divide \cite{skp+15} & {\scriptsize$R_{50}^{128}$} &
	- &
	- &
	- &
	- &
	
	$75.9$ & 
	$88.4$ & 
	$94.9$ &
	$98.1$ &

    $87.7$ &
    $92.9$ &
    $85.7$ &
    $90.4$ &
    $82.9$ &
    $90.2$ &
    
    - &
    - &
    - &
    -
	\\
	
	MIC \cite{rbo+19} & {\scriptsize$R_{50}^{128}$} &
	- &
	- &
	- &
	- &

	$77.2$ &
	$89.4$ &
	$95.6$ &
	-  &

    $86.9$ &
    $93.4$ &
    - &
    - &
    $82.0$ &
    $91.0$ &
    
    - &
    - &
    - &
    -

	\\
	
	Cont. w/M \cite{wzh+20} & {\scriptsize$R_{50}^{128}$} &
	- &
	- &
	- &
	- &
	
	$\underline{80.6}$ &
	$\underline{91.6}$ &
	$\underline{96.2}$ &
	$\underline{98.7}$ &

    $\underline{94.7}$ &
    $\underline{96.8}$ &
    $\underline{93.7}$ &
    $\underline{95.8}$ &
    $\underline{93.0}$ &
    $\underline{95.8}$ &
    
    - &
    - &
    - &
    -
	\\
	
	\hline
	RS@k\textsuperscript{\dag} & {\scriptsize$R_{50}^{128}$} &
	$69.3$  &
	$82.9$  &
	$90.6$  &
	$93.1$ &
	
	$80.6$ &
	$91.6$ &
	$96.4$ &
	$\bm{98.8}$ &

    $\bm{95.6}$ &
    $\bm{97.8}$ &
    $\bm{94.4}$ &
    $\bm{96.8}$ &
    $\bm{93.5}$ &
    $\bm{96.6}$ &
    
    $78.1$ &
    $85.8$ &
    $91.1$ &
    $94.5$
	\\

%
%
%
	
	RS@k\textsuperscript{\dag} +SiMix & {\scriptsize$R_{50}^{128}$} &
	$\bm{69.6}$ & 
	$\bm{83.3}$ & 
	$\bm{91.2}$ & 
	$\bm{93.8}$ &
	
	$\bm{80.9}$ &
	$\bm{91.7}$ &
	$\bm{96.5}$ &
	$\bm{98.8}$ &

    $95.4$ &
    $97.5$ &
    $93.8$ &
    $96.6$ &
    $93.0$ &
    $96.2$ &
    
    $\bm{84.7}$ &
    $\bm{90.9}$ &
    $\bm{94.7}$ &
    $\bm{96.9}$ 
	\\

	



%
%
%

    \hline \hline
	FastAP \cite{chx+19} & {\scriptsize$R_{50}^{512}$} &
	$60.6$ & 
	$77.0$ & 
	$87.2$ & 
	$90.6$ &

	$76.4$ &
	$89.0$ & 
	$95.1$ & 
	$98.2$ &

    $91.9$ &
    $96.8$ &
    $90.6$ &
    $95.9$ &
    $87.5$ &
    $95.1$ &
    
    - &
    - &
    - &
    -
	\\
	
	MS \cite{whh+19} & {\scriptsize$I_{3}^{512}$} &
	- &
	- &
	- &
	- &
	
	$78.2$ &
	$90.5$ &
	$96.0$ &
	$98.7$ &

	- &
	- &
	- &
	- &
	- &
	- &
	
	$84.1$ &
	$90.4$ &
	$94.0$ &
	$96.1$
	\\
	
	NormSoftMax \cite{zw18} & {\scriptsize$R_{50}^{512}$} &
	- &
	- &
	- &
	- &
	
	$78.2$ &
	$90.6$ &
	$96.2$ &
	- &

	- &
	- &
	- &
	- &
	- &
	- &
	
	$84.2$ &
	$90.4$ &
	$94.4$ &
	$96.9$
	\\
	
	Blackbox AP \cite{rmp+20} & {\scriptsize$R_{50}^{512}$} &
	$62.9$ &
	$79.0$ &
	$88.9$ &
	$92.1$ &
	
	$78.6$ &
	$90.5$ &
	$96.0$ &
	$98.7$ &
	
	- &
	- &
	- &
	- &
	- &
	- &

    - &
    - &
    - &
    -
	\\

	ROADMAP \cite{ramzi2021robust} & {\scriptsize$R_{50}^{512}$} &
	$69.1$ &
	$83.1$ &
	$91.3$ &
	$93.9$ &

	$\bm{83.1}$ &
	$\underline{92.7}$ &
	$96.3$ &
	-      &

    - &
    - &
    - &
    - &
    - &
    - &
    
    - &
    - &
    - &
    -
	\\	

	IntraBatch \cite{sel21} & {\scriptsize$R_{50}^{512}$} &
	- &
	- &
	- &
	- &

	$81.4$ &
	$91.3$ &
	$95.9$ &
	-      &

    - &
    - &
    - &
    - &
    - &
    - &
    
    $88.1$ &
    $\bm{93.3}$ &
    $\bm{96.2}$ &
    $\bm{98.2}$
	\\	

	CIA \cite{kpm+23} & {\scriptsize$R_{50}^{512}$} &
	- &
	- &
	- &
	- &

	$82.3$ &
	- &
	- &
	-      &

    - &
    - &
    - &
    - &
    - &
    - &
    
    $\bm{91.2}$ &
    - &
    - &
    -
	\\

	Cont. w/M \cite{wzh+20} & {\scriptsize$I_{3}^{512}$} &
	- &
	- &
	- &
	- &
	
	$79.5$ &
	$90.8$ &
	$96.1$ &
	$98.7$ &

    $94.6$ &
    $96.9$ &
    $\underline{93.4}$ &
    $96.0$ &
    $\underline{93.0}$ &
    $96.1$ &

    - &
    - &
    - &
    -
	\\
	
	HORDE \cite{jph+19} & {\scriptsize$R_{50}^{512}$} &
	- &
	- &
	- &
	- &
	
	$80.1$ &
	$91.3$ &
	$96.2$ &
	- &

	- &
	- &
	- &
	- &
	- &
	- &
	
	$86.2$ &
	$91.9$ &
	$95.1$ &
	$97.2$
	\\
	
	ProxyNCA++ \cite{tdt20} & {\scriptsize$R_{50}^{512}$} &
	- &
	- &
	- &
	- &
	
	$80.7$ &
	$92.5$ &
	$96.7$ &
	$98.9$ &

	- &
	- &
	- &
	- &
	- &
	- &
	
	$86.5$ &
	$92.5$ &
	$95.7$ &
	$\underline{97.7}$
	\\

	SAP \cite{bxk+20} & {\scriptsize$R_{50}^{512}$} &
	$\underline{67.2}$ &
	$\underline{81.8}$ &
	$\underline{90.3}$ &
	$\underline{93.1}$ &
	
	$80.1$ &
	$91.5$ &
	$\underline{96.6}$ & 
	$\underline{99.0}$ &

    $\underline{94.9}$ &
    $\underline{97.6}$ &
    $93.3$ &
    $\underline{96.4}$ &
    $91.9$ &
    $\underline{96.2}$ &
    
    $76.1$ &
    $84.3$ &
    $89.8$ &
    $93.8$
	\\
	
	SAP\textsuperscript{\dag} \cite{bxk+20} {\tiny +GeM +LN} & {\scriptsize$R_{50}^{512}$} &
	$68.7$ &
	$82.7$ &
	$90.9$ &
	$93.5$ &
	
	$80.3$ &
	$92.0$ &
	$96.9$ &
	$99.0$ &

    $94.2$ &
    $97.2$ &
    $92.7$ &
    $96.2$ &
    $91.0$ &
    $95.8$ &
    
    $78.2$ &
    $85.6$ &
    $90.8$ &
    $94.3$
	\\

	

    

	
	
	

	
    
	

	\hline
	
	

	RS@k\textsuperscript{\dag} & {\scriptsize$R_{50}^{512}$} &
	$71.2$ &
	$84.0$ &
	$91.3$ &
	$93.6$ &
	
	$\underline{82.8}$ & 
	$\bm{92.9}$ & 
	$\bm{97.0}$ & 
	$99.0$ &
	
    $\bm{95.7}$ &
    $\bm{97.9}$ &
    $\bm{94.6}$ &
    $\bm{96.9}$ &
    $\bm{93.8}$ &
    $\bm{96.6}$ &
    
    $80.7$ &
    $88.3$ &
    $92.8$ &
    $95.7$
	\\

%
%
%

	RS@k\textsuperscript{\dag} +SiMix & {\scriptsize$R_{50}^{512}$} &
	$\bm{71.8}$ &
	$\bm{84.7}$ &
	$\bm{91.9}$ &
	$\bm{94.3}$ &
	
	$82.1$ & 
	$92.8$ &
	$\bm{97.0}$ &
	$\bm{99.1}$ &

    $95.3$ &
    $97.7$ &
    $94.2$ &
    $96.5$ &
    $93.3$ &
    $96.4$ &
    
    $\underline{88.2}$ &
    $\underline{93.0}$ &
    $\underline{95.9}$ &
    $97.4$
	\\
	
	



%
%
%
%

    \hline \hline
    
    SAP\textsuperscript{\dag} \cite{bxk+20} & {\scriptsize ViT-B/32$^{512}$} &
    $72.2$ &
	$84.6$ &
	$91.6$ &
	$93.9$ &
	
    $83.7$ &
	$94.0$ &
	$97.8$ &
	$99.3$ &
	
	$94.8$ &
	$97.7$ &
	$93.5$ &
	$96.8$ &
	$92.1$ &
	$96.3$ &
	
	$78.1$ &
	$85.7$ &
	$91.0$ &
	$94.8$
	\\

    RS@k\textsuperscript{\dag} & {\scriptsize ViT-B/32$^{512}$} & 
    $75.9$ & 
    $87.1$ &
    $93.1$ &
    $95.1$ &

    $85.1$ & 
    $94.6$ &
    $98.0$ &
    $99.3$ &

    $95.1$ & 
    $97.7$ &
    $94.1$ &
    $96.7$ &
    $93.2$ &
    $96.5$ &
    
    $78.1$ &
    $86.4$ &
    $92.3$ &
    $95.6$
    \\

%
%
    
	\hline
	
    SAP\textsuperscript{\dag} \cite{bxk+20} & {\scriptsize ViT-B/16$^{512}$} &
	$79.1$ &
	$89.0$ &
	$94.2$ &
	$95.8$ &
	
	$86.6$ &
	$95.4$ &
	$98.4$ &
	$99.5$ &

	$95.5$ &
	$97.7$ &
	$94.2$ &
	$96.9$ &
	$93.1$ &
	$96.6$ &
	
	$86.2$ &
	$92.1$ &
	$95.1$ &
	$97.2$
	\\

    RS@k\textsuperscript{\dag} & {\scriptsize ViT-B/16$^{512}$} & 
    $83.9$ & 
    $92.1$ &
    $95.9$ &
    $97.2$ &
    
    $88.0$ & 
    $96.1$ &
    $98.6$ &
    $99.6$ &
    
    $96.2$ & 
    $98.0$ &
    $95.2$ &
    $97.2$ &
    $94.7$ &
    $97.1$ &
    
    $89.5$ &
    $94.2$ &
    $96.6$ &
    $98.3$
    \\

%
%
    \hline
    \end{tabular}

%% file: tab_metriclearning_initialization.tex
\begin{tabular}{lll@{\lsp}cccc@{\lsp}cccc@{\lsp}cccc}
\hline
 \multirow{1}[1]{*}{Init.} & \multirow{1}[1]{*}{Arch.$^\text{dim}$} & \multirow{1}[1]{*}{Method}  & \multicolumn{4}{c}{iNaturalist \cite{vms+18}} & \multicolumn{4}{c}{SOP \cite{ohb16}}  & \multicolumn{4}{c}{Cars196 \cite{ksd+13}} \\
& & & $1$ & $4$ & $16$ & $32$ & $10^0$ & $10^1$ & $10^2$ & $10^3$ & $1$ & $2$ & $4$ & $8$ \\
\hline
\multirow{9}[3]{*}{IN-21K} & \multirow{3}[1]{*}{ViT-B/$32^{512}$}  & init &
$46.9$ & 
$64.1$ & 
$78.2$ & 
$83.7$ & 

$54.7$ & 
$69.9$ & 
$82.9$ & 
$93.2$ &

$33.2$ & 
$44.1$ & 
$55.8$ & 
$68.1$ \\



& & SAP & 

$72.2$ &
$84.6$ &
$91.6$ &
$93.9$ &

$83.7$ &
$94.0$ &
$97.8$ &
$99.3$ &

$78.1$ &
$85.7$ &
$91.0$ &
$94.8$ \\


& &  RS@k & 

$75.9$ & 
$87.1$ &
$93.1$ &
$95.1$ &

$85.1$ & 
$94.6$ &
$98.0$ &
$99.3$ &

$78.1$ &
$86.4$ &
$92.3$ &
$95.6$ \\


\cdashline{4-15}

& \multirow{3}[1]{*}{ViT-B/$16^{512}$} & -  &
$59.3$ & 
$74.9$ & 
$85.9$ & 
$89.8$ &

$55.8$ & 
$70.9$ & 
$83.4$ & 
$93.6$ &

$35.6$ & 
$46.9$ & 
$58.6$ & 
$70.5$ \\


& & SAP & 

$79.1$ &
$89.0$ &
$94.2$ &
$95.8$ &

$86.6$ &
$95.4$ &
$98.4$ &
$99.5$ &

$86.2$ &
$92.1$ &
$95.1$ &
$97.2$ \\


&  & RS@k &

$83.9$ & 
$92.1$ &
$95.9$ &
$97.2$ &

$88.0$ & 
$96.1$ &
$98.6$ &
$99.6$ &

$89.5$ &
$94.2$ &
$96.6$ &
$98.3$ \\


\cdashline{4-15}

& \multirow{3}[1]{*}{ViT-L/$16^{512}$} & init &
$69.3$ & 
$83.2$ & 
$91.4$ & 
$94.1$ &

$64.9$ & 
$79.4$ & 
$89.4$ & 
$96.5$ &

$50.1$ & 
$62.1$ & 
$73.0$ & 
$82.8$ \\


& &SAP &

$82.6$ & 
$91.1$ & 
$95.4$ & 
$96.7$ & 

$86.6$ &
$95.7$ &
$98.5$ &
$99.6$ &

$91.3$ & 
$95.0$ & 
$97.3$ & 
$98.5$ \\


&  & RS@k & 

$84.9$ & 
$92.4$ & 
$96.0$ & 
$97.2$ & 

$88.6$ & 
$96.6$ & 
$98.9$ & 
$99.6$ & 

$90.4$ & 
$94.4$ & 
$96.8$ & 
$98.2$ \\


\hline

\multirow{9}[3]{*}{CLIP-400M} & \multirow{3}[1]{*}{ViT-B/$32^{512}$}  & -  &
$53.9$ & 
$72.1$ & 
$84.7$ & 
$89.2$ &

$62.7$ & 
$78.6$ & 
$89.5$ & 
$96.7$ &

$87.0$ & 
$93.1$ & 
$96.3$ & 
$98.2$ \\


& & SAP & 

$62.6$ &
$78.1$ &
$87.6$ &
$90.8$ &

$80.6$ &
$92.5$ &
$97.3$ &
$99.3$ &

$92.6$ &
$96.1$ &
$97.8$ &
$98.8$ \\


& &  RS@k & 

$70.4$ &
$83.7$ &
$91.1$ &
$93.7$ &

$84.3$ &
$94.8$ &
$98.1$ &
$99.5$ &

$91.8$ &
$95.7$ &
$97.9$ &
$98.8$ \\


\cdashline{4-15}

& \multirow{3}[1]{*}{ViT-B/$16^{512}$} & init &
$62.9$ & 
$79.7$ & 
$89.8$ & 
$93.1$ & 

$65.7$ & 
$81.3$ & 
$91.2$ & 
$97.2$ &

$91.4$ & 
$95.5$ & 
$97.8$ & 
$98.8$\\


& & SAP & 

$73.7$ &
$85.7$ &
$92.3$ &
$94.5$ &

$85.0$ &
$94.9$ &
$98.2$ &
$99.6$ &

$95.2$ &
$97.4$ &
$98.5$ &
$99.0$ \\


&  & RS@k &

$80.1$ &
$89.9$ &
$94.9$ &
$96.4$ &

$87.7$ &
$96.2$ &
$98.7$ &
$99.7$ &

$95.5$ &
$97.5$ &
$98.5$ &
$99.1$ \\


\cdashline{4-15}
& \multirow{3}[1]{*}{ViT-L/$14^{512}$} & -  &
$70.1$ & 
$84.6$ & 
$92.6$ & 
$95.0$ &

$67.2$ & 
$82.4$ & 
$91.8$ & 
$97.4$ &

$93.4$ & 
$97.0$ & 
$98.5$ & 
$99.3$ \\


& & SAP &

$77.8$ &
$89.0$ &
$94.6$ &
$96.2$ &

$86.2$ &
$95.5$ &
$98.5$ &
$99.6$ &

$96.6$ &
$98.1$ &
$\underline{98.8}$ &
$\underline{99.2}$ \\


&  & RS@k & 

$82.6$ &
$91.4$ &
$95.6$ &
$96.9$ &

$89.3$ &
$97.1$ &
$99.1$ &
$99.6$ &

$96.7$ &
$98.1$ &
$\underline{98.8}$ &
$\underline{99.2}$ \\


\hline

\multirow{9}[3]{*}{CLIP-2B} & \multirow{3}[1]{*}{ViT-B/$32^{512}$}  & init &
$56.2$ & 
$74.6$ & 
$86.8$ & 
$90.8$ &

$64.1$ & 
$80.0$ & 
$90.5$ & 
$96.8$ &

$90.1$ & 
$95.1$ & 
$97.6$ & 
$98.9$ \\


& & SAP & 

$62.9$ & 
$78.7$ & 
$88.5$ &
$91.8$ &

$82.3$ & 
$93.5$ & 
$97.8$ & 
$99.5$ & 

$94.4$ & 
$97.0$ & 
$98.1$ & 
$99.0$ \\ 


& &  RS@k & 

$71.2$ & 
$84.4$ & 
$91.7$ & 
$94.0$ &

$86.4$ & 
$95.7$ & 
$98.6$ & 
$99.7$ & 

$94.2$ & 
$97.0$ & 
$98.4$ & 
$99.1$ \\ 


\cdashline{4-15}

& \multirow{3}[1]{*}{ViT-B/$16^{512}$} & init &
$64.4$ & 
$80.8$ & 
$90.6$ & 
$93.6$ &

$67.0$ & 
$82.4$ & 
$91.8$ & 
$97.3$ &

$92.7$ & 
$96.7$ & 
$98.4$ & 
$99.3$ \\


& & SAP & 

$74.2$ & 
$86.2$ & 
$92.6$ & 
$94.7$ &

$86.6$ & 
$95.7$ & 
$98.6$ & 
$99.7$ & 

$96.5$ & 
$98.0$ & 
$98.7$ & 
$\underline{99.2}$ \\ 


&  & RS@k &

$80.3$ & 
$90.2$ & 
$95.0$ & 
$96.5$ &

$89.0$ & 
$96.9$ & 
$99.1$ & 
$\underline{99.8}$ & 

$96.4$ & 
$98.1$ & 
$\underline{98.8}$ & 
$\underline{99.2}$ \\ 


\cdashline{4-15}

& \multirow{3}[1]{*}{ViT-L/$14^{512}$} & init &
$66.0$ & 
$82.0$ & 
$91.3$ & 
$94.2$ &

$65.9$ & 
$81.2$ & 
$91.1$ & 
$97.1$ &

$93.8$ & 
$97.3$ & 
$98.7$ & 
$99.4$ \\


& & SAP &

$77.9$ & 
$89.1$ & 
$94.6$ & 
$96.2$ & 

$87.5$ & 
$96.1$ & 
$98.8$ & 
$99.7$ & 

$96.9$ & 
$98.3$ & 
$\underline{98.8}$ & 
$\underline{99.2}$ \\ 


&  & RS@k & 

$82.8$ & 
$91.5$ & 
$95.8$ & 
$97.0$ &

$\underline{90.6}$ & 
$\underline{97.5}$ & 
$\underline{99.3}$ & 
$\bm{99.9}$ & 

$97.0$ & 
$98.2$ & 
$\underline{98.8}$ & 
$\underline{99.2}$ \\ 


\hline

\multirow{9}[3]{*}{DiHT} & \multirow{3}[1]{*}{ViT-B/$32^{512}$}  & init &
$58.5$ & 
$76.3$ & 
$87.8$ & 
$91.6$ &

$62.2$ & 
$78.5$ & 
$89.6$ & 
$97.0$ &

$87.7$ & 
$93.9$ & 
$97.1$ & 
$98.7$ \\


& & SAP & 

$66.8$ & 
$81.1$ & 
$89.5$ & 
$92.3$ & 

$83.3$ & 
$94.0$ & 
$98.0$ & 
$99.5$ & 

$93.6$ & 
$96.6$ & 
$98.1$ & 
$98.9$ \\ 


& &  RS@k & 

$73.3$ & 
$85.6$ & 
$92.4$ & 
$94.6$ & 

$85.9$ & 
$95.5$ & 
$98.5$ & 
$99.7$ & 

$92.8$ & 
$96.4$ & 
$98.1$ & 
$99.1$ \\


\cdashline{4-15}

& \multirow{3}[1]{*}{ViT-B/$16^{512}$} & init &
$67.2$ & 
$82.9$ & 
$91.8$ & 
$94.6$ & 

$65.6$ & 
$81.1$ & 
$91.4$ & 
$97.6$ &

$91.1$ & 
$95.6$ & 
$97.9$ & 
$98.9$ \\


& & SAP & 

$76.8$ & 
$87.9$ & 
$93.6$ & 
$95.4$ & 

$87.0$ & 
$96.0$ & 
$98.8$ & 
$99.7$ & 

$96.1$ & 
$97.8$ & 
$98.7$ & 
$\underline{99.2}$ \\ 


&  & RS@k &

$82.4$ & 
$91.5$ & 
$95.8$ & 
$97.1$ & 

$88.9$ & 
$96.9$ & 
$99.1$ & 
$\underline{99.8}$ & 

$96.2$ & 
$97.9$ & 
$98.6$ & 
$\underline{99.2}$ \\ 


\cdashline{4-15}
& \multirow{3}[1]{*}{ViT-L/$14^{512}$} & init &
$74.9$ & 
$87.8$ & 
$94.6$ & 
$96.5$ & 

$68.3$ & 
$83.4$ & 
$92.9$ & 
$98.1$ &

$93.9$ & 
$97.1$ & 
$98.6$ & 
$99.4$ \\


& & SAP &

$81.3$ & 
$91.1$ & 
$95.7$ & 
$97.0$ & 

$87.4$ & 
$96.0$ & 
$98.8$ & 
$\underline{99.8}$ & 

$\underline{97.1}$ & 
$98.4$ & 
$\bm{99.0}$ & 
$\bm{99.3}$ \\ 


&  & RS@k & 

$85.2$ & 
$92.9$ & 
$96.5$ & 
$97.5$ & 

$\bm{90.8}$ & 
$\bm{97.7}$ & 
$\bm{99.4}$ & 
$\underline{99.8}$ & 

$\bm{97.2}$ & 
$\bm{98.5}$ & 
$\bm{99.0}$ & 
$\bm{99.3}$\\ 


\hline

\multirow{6}[2]{*}{SWAG} & \multirow{3}[1]{*}{ViT-B/$16^{512}$} & init &
$74.9$ & 
$86.9$ & 
$93.5$ & 
$95.6$ & 

$57.5$ & 
$73.6$ & 
$86.4$ & 
$95.5$ &

$76.5$ & 
$85.7$ & 
$91.5$ & 
$95.6$ \\


& & SAP & 

$80.4$ & 
$90.3$ & 
$95.1$ & 
$96.5$ & 

$85.8$ & 
$95.1$ & 
$98.3$ & 
$99.6$ & 

$92.8$ & 
$96.0$ & 
$97.7$ & 
$98.7$ \\ 


&  & RS@k &

$81.8$ & 
$91.0$ & 
$95.4$ & 
$96.8$ & 

$88.1$ & 
$96.3$ & 
$98.8$ & 
$99.7$ & 

$92.5$ & 
$95.9$ & 
$97.5$ & 
$98.5$ \\ 


\cdashline{4-15}

& \multirow{3}[1]{*}{ViT-L/$16^{512}$} & init &
$73.1$ & 
$86.1$ & 
$93.3$ & 
$95.5$ &

$59.3$ & 
$75.2$ & 
$87.3$ & 
$96.1$ &

$81.1$ & 
$88.8$ & 
$94.2$ & 
$97.5$ \\


& & SAP &

$83.5$ & 
$92.0$ & 
$95.9$ & 
$97.1$ & 

$85.8$ & 
$95.0$ & 
$98.2$ & 
$99.5$ & 

$95.0$ & 
$97.3$ & 
$98.2$ & 
$98.9$ \\ 


&  & RS@k & 

$84.9$ & 
$92.6$ & 
$96.2$ & 
$97.3$ & 

$89.2$ & 
$96.7$ & 
$98.9$ & 
$99.7$ & 

$94.8$ & 
$97.2$ & 
$98.2$ & 
$99.1$ \\


\hline

\multirow{3}[1]{*}{DINOv2} & \multirow{3}[1]{*}{ViT-L/$14^{512}$} & init &
$82.2$ & 
$90.8$ & 
$95.3$ & 
$96.7$ &

$55.6$ & 
$71.4$ & 
$83.8$ & 
$95.0$ &

$82.9$ & 
$88.3$ & 
$92.1$ & 
$95.1$ \\


& & SAP &

$\underline{88.5}$ & 
$\underline{94.7}$ & 
$\underline{97.5}$ & 
$\underline{98.2}$ & 

$87.5$ & 
$96.1$ & 
$98.7$ & 
$99.7$ & 

$96.3$ & 
$97.9$ & 
$98.6$ & 
$99.0$ \\ 


&  & RS@k & 

$\bm{90.0}$ & 
$\bm{95.4}$ & 
$\bm{97.6}$ & 
$\bm{98.3}$ & 

$90.5$ & 
$\underline{97.5}$ & 
$99.2$ & 
$\underline{99.8}$ & 

$96.8$ & 
$98.2$ & 
$98.6$ & 
$99.1$ \\ 


\hline

\end{tabular}

%% file: 4_conclusions.tex
\section{Conclusions}
\label{sec:conclusions}
This work presents a novel surrogate loss function for the recall@k metric to perform trainining for deep metric learning. 
The need for large batch size is pronounced and faciliated via two ways, an implementation framework that dispenses with the hardware constraints to a good extend and the proposed SimMix that efficiently and virtually increases the batch size. 
State-of-the-art results are achieved on a number of standard benchmarks for deep metric learning and instance-level search with different backbones. 
Using model initialization from recent large-scale pre-training approaches, further boosts the results to a level that some of the benchmarks seem nearly solved. 